\documentclass[twocolumn]{fairmeta}

\usepackage[most]{tcolorbox}
\usepackage[table]{xcolor}
\usepackage{algorithm}
\usepackage{algorithmic}
\definecolor{ntuRed}{HTML}{D71920}
\definecolor{ntuBlue}{HTML}{003D7C}
\definecolor{ntuBest}{RGB}{255,235,238}
\definecolor{ntuSecond}{RGB}{235,245,255}
\definecolor{ntuBaseBest}{RGB}{255,244,214}

\usepackage{booktabs}
\usepackage{amsmath}
\usepackage{amssymb}
\usepackage{pifont}
\usepackage{float}
\usepackage{placeins}

\newcommand{\best}[1]{\cellcolor{ntuBest}\textbf{#1}}
\newcommand{\second}[1]{\cellcolor{ntuSecond}#1}

\newcommand{\gcmark}{\textcolor{green!70!black}{\ding{51}}}
\newcommand{\rxmark}{\textcolor{red}{\ding{55}}}

\usepackage{hyperref}
\definecolor{appPurple}{RGB}{105,45,150}

\newcommand{\appentry}[3]{%
  \noindent{\color{ntuBlue}\large\bfseries #1\quad #2}
  \dotfill
  {\bfseries \pageref{#3}}\par\vspace{0.75em}
}

\newcommand{\appsubentry}[3]{%
  \noindent\hspace{2.8em}{\color{ntuBlue}#1\quad #2}
  \dotfill
  \pageref{#3}\par\vspace{0.45em}
}

\newcommand{\appheading}[2]{%
  \clearpage
  \phantomsection
  \label{#1}
  \noindent{\Large\bfseries #2}\par
  \vspace{0.8em}
}

\newcommand{\appsubheading}[2]{%
  \phantomsection
  \label{#1}
  \vspace{1.0em}
  \noindent{\large\bfseries #2}\par
  \vspace{0.5em}
}

\title{$\omega$-0: A Latent Predictive World Action Model for Concurrent Humanoid Loco-Manipulation}

\author[1, 3, *, \dagger]{Zhe Li}
\author[2, 3, *]{Zhenzhe Zhang}
\author[3, *]{Yangyang Wei}
\author[4, *]{Wenjie Zhang}
\author[1, *]{Xichen Yuan}
\author[3]{Peiyuan Zhi}
\author[1]{Gen Li}
\author[1]{Xinying Guo}
\author[1]{Fengjie Gao}
\author[1, \clubsuit]{Jianfei Yang}
\author[2, \clubsuit]{Shanghang Zhang}

\affiliation[1]{MARS Lab, NTU}
\affiliation[2]{PKU}
\affiliation[3]{BAAI}
\affiliation[4]{HKUST(GZ)}

\contribution[*]{Equal Contribution}
\contribution[\dagger]{Project Lead}
\contribution[\clubsuit]{Corresponding Authors}

\correspondence{Shanghang Zhang at \email{shanghang@pku.edu.cn}; Jianfei Yang at \email{jianfei.yang@ntu.edu.sg}. Project Page in \email{https://gentlefress.github.io/OMEGA-0\_page/}}

\abstract{
Humanoid household tasks often require concurrent loco-manipulation, where the robot must move, adjust posture, maintain balance, and manipulate objects as a single coordinated behavior. 
Yet existing humanoid policies typically decompose locomotion and manipulation, while recent world-action models remain either arm-centric or video-centered. 
We present $\omega$-0, a latent predictive whole-body world-action model for real-world humanoid concurrent loco-manipulation. 
Given a language instruction, current visual observation, and robot proprioceptive state, $\omega$-0 directly predicts controller-compatible whole-body action latents for real-robot execution. 
Rather than reconstructing future videos, $\omega$-0 learns compact future observation embeddings as a lightweight predictive objective, coupling latent visual foresight with diffusion-based whole-body action generation. 
The model supports egocentric RGB, exocentric RGB, and exocentric depth inputs, and leverages controller-based simulation replay to ground human/public visual-motion priors into robot-executable action latents. 
We further collect $\omega$-HOME, a 40+ hour real-world household humanoid dataset with synchronized multi-view observations, whole-body SMPL motions, robot states, and action latents. 
Real-world experiments on 11 household tasks demonstrate that a single $\omega$-0 model can produce smooth manipulate-while-moving behaviors and consistently outperform representative imitation learning, VLA, humanoid, and WAM baselines.
}

\begin{document}

\teaser{
\begin{center}
    \includegraphics[width=0.9\textwidth]{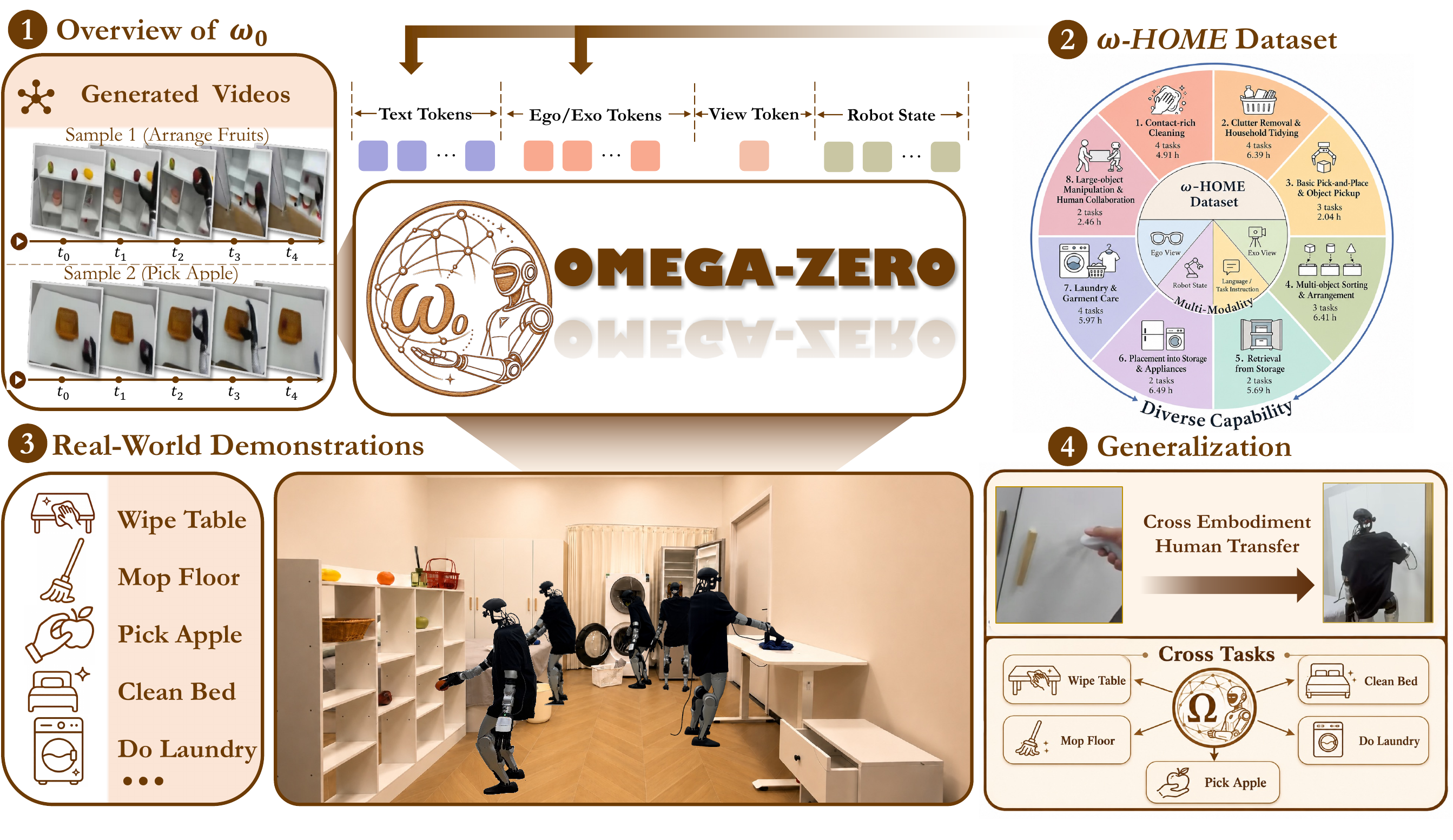}
    \captionof{figure}{
    Overview of $\omega$-0 and the $\omega$-HOME dataset.
    $\omega$-0 generates whole-body action latents from language, multi-view observations, and robot states, while $\omega$-HOME provides 40h of multimodal household humanoid demonstrations for training and evaluation.
    }
    \label{fig:motivation}
\end{center}
}

\maketitle

\section{Introduction}

Humanoid robots are attractive for household environments because their bodies are compatible with human-centered spaces, furniture, and tools. 
However, household assistance requires more than arm manipulation or navigation alone. 
A robot wiping a large table must step, lean, and maintain contact with the surface; a robot mopping the floor must move its base while controlling a long tool; a robot retrieving objects from a refrigerator or placing clothes into a lower washing machine compartment must coordinate reaching, bending, balance, and hand interaction. 
These tasks are not simply long sequences of independent skills. 
They require concurrent loco-manipulation, where the lower body, torso, arms, and hands continuously adapt to one another while the robot manipulates during movement.

This setting exposes a gap in current robot learning systems. 
Many vision-language-action policies are designed around arm-centric manipulation, where the policy predicts end-effector commands or arm actions from visual observations and language instructions~\citep{chi2025diffusion, intelligence2025pi, zitkovich2023rt, brohan2022rt}. 
Mobile manipulation systems extend the workspace with a wheeled base~\citep{yenamandra2023homerobot, wu2023tidybot, wu2024tidybot++, fu2024mobile}, but the base and arms are still often treated as separate functional components. 
Recent humanoid VLAs make important progress toward loco-manipulation~\citep{wei2026psi_0, hu2026openhlm, jiang2025wholebodyvla, yang2025egovla, gr00tn1_2025}, yet many systems still rely on practical decompositions between locomotion, balance, and manipulation. 
Such decompositions can be effective when the robot first moves to a location and then manipulates while mostly standing still. 
However, they become limiting in tasks where stable execution depends on simultaneous stepping, torso adjustment, reaching, and contact maintenance. 
This motivates a policy representation in which whole-body coordination is learned directly, rather than emerging indirectly from separately designed modules.

World action models (WAMs)~\citep{hu2024video, yuan2026fast, pai2025mimic, cen2025worldvla, li2025unified, li2026causal, kim2026cosmos, ye2026gigaworld, ye2026world, bi2025motus} offer a natural way to learn such coordinated behaviors. 
Instead of predicting actions only from the current observation, WAMs use future visual dynamics to provide additional supervision about task progress and the consequences of actions. 
This is especially relevant for humanoid household tasks: whether the robot is wiping effectively, moving toward the correct object, keeping a tool in contact, or approaching a receptacle is often reflected in how the scene evolves over time. 
However, most existing WAMs are developed for arm-centric manipulation, where future prediction primarily supports local object interaction. 
Recent humanoid WAMs~\citep{zheng2026motionwam} begin to extend this idea to whole-body control, but they often make video dynamics a central intermediate representation for action prediction. 
In real-world humanoid tasks, visual observations are noisy, occluded by the robot body or tools, and affected by viewpoint changes during locomotion. 
When action generation depends strongly on a predicted video trajectory, temporal inconsistencies in that trajectory can be amplified into abrupt transitions, hesitant motion, or unstable whole-body coordination. 
Moreover, improving pixel-level video fidelity does not necessarily translate into better control: for real-time humanoid execution, the policy mainly needs compact future information that is useful for choosing the next whole-body action.

We therefore ask a different question: \textbf{can future prediction be used not as a video generation target, but as a compact predictive signal for whole-body action learning?} 
Our design follows this principle.
Rather than building a video-centered pipeline that first predicts future dynamics and then converts them into actions, we learn a latent predictive world-action representation in which future observation embeddings and whole-body action latents are trained together. 
Future prediction provides task-progress and scene-evolution cues, while the action branch directly generates controller-compatible whole-body action latents. 
This design preserves the benefit of world modeling, but avoids relying on large video generators or test-time video-to-action inversion.

We propose $\omega$-0, a latent predictive world-action model for humanoid concurrent loco-manipulation. 
Given a language instruction, current visual observation, and robot proprioceptive state, $\omega$-0 predicts a future chunk of whole-body action latents that are executed by a low-level whole-body controller~\citep{luo2025sonic}. 
In parallel, a lightweight future visual latent branch predicts compact future observation embeddings as an auxiliary predictive objective. 
This joint-embedding predictive formulation couples latent visual foresight with whole-body action generation, enabling the policy to produce coordinated manipulate-while-moving behaviors without decomposing execution into separate locomotion and manipulation phases.

A second challenge is the scarcity of real-world humanoid data.
Real-world humanoid demonstrations are costly to collect, while human demonstrations and public video-action datasets contain rich visual-motion priors. 
However, these data cannot be directly used as humanoid policies because human motion, robot state, and controller actions lie in different representation spaces. 
To bridge this gap, $\omega$-0 first learns action-aware visual-language representations from whole-body action tokens, and then grounds human/public visual-motion priors into robot-executable action latents through SONIC-based simulation replay. 
Motions that cannot be reliably tracked are filtered out, and the remaining trajectories provide controller-compatible action latents and proprioceptive supervision. 
This allows scalable human data to contribute to humanoid action learning while preserving real-robot executability.

The same latent formulation also supports flexible visual conditioning. 
$\omega$-0 can take egocentric RGB, exocentric RGB, or exocentric depth observations as input, with view tokens distinguishing different camera perspectives. 
Egocentric observations provide deployment-time visual feedback, while exocentric RGB-D observations offer complementary supervision about whole-body motion, object-scene relations, and task progress during training. 
This enables the model to benefit from informative third-person observations while still supporting first-person execution on the robot.

To train and evaluate $\omega$-0, we collect $\omega$-HOME, a 40-hour real-world household humanoid dataset containing synchronized egocentric RGB videos, exocentric RGB videos, exocentric depth videos, whole-body SMPL motions~\citep{loper2023smpl}, robot proprioceptive states, and whole-body action latents. 
We evaluate $\omega$-0 on 11 household loco-manipulation tasks covering tabletop manipulation, cleaning, laundry handling, object transfer, appliance interaction, and mobile manipulation. 
Importantly, all tasks are executed by a single unified $\omega$-0 model, rather than by task-specific policies, task-specific action heads, or separate locomotion and manipulation modules. 
Real-world rollouts show smooth whole-body coordination during simultaneous stepping, reaching, contact-rich manipulation, and object interaction. 
We further observe promising generalization to held-out objects, scenes, and human-data transfer settings.

Our contributions are summarized as follows:
\begin{itemize}
\item We introduce $\omega$-0, a latent predictive whole-body humanoid world-action model that couples future visual embedding prediction with diffusion-based whole-body action generation for concurrent loco-manipulation.

\item We develop a staged training pipeline that learns action-aware visual-language representations and grounds human/public visual-motion priors into robot-executable action latents through SONIC-based simulation replay and real-world post-training.

\item We collect $\omega$-HOME, a 40-hour real-world household humanoid dataset with synchronized multi-view RGB-D observations, whole-body SMPL motions, robot states, and action latents, providing multimodal supervision for whole-body humanoid loco-manipulation.

\item We demonstrate that a single $\omega$-0 model can autonomously execute 11 diverse real-world household tasks, covering both short-horizon and long-horizon loco-manipulation, without training task-specific policies.
\end{itemize}
\section{Related Work}
\subsection{Whole-Body Humanoid Control}
Physics-based whole-body control aims to generate dynamically feasible humanoid motions from reference trajectories while maintaining balance, contact consistency, and temporal smoothness. 
Early works~\citep{chen2025gmt, ji2024exbody2, xie2026kungfubot, he2025asap, li2025language, li2025robomirror, li2025you} establish reinforcement-learning-based motion tracking as a standard paradigm for humanoid control, and later methods improve robustness for real-world deployment across different humanoid platforms. 
Recent approaches further scale whole-body tracking to broader motion distributions and more challenging behaviors. 
GMT~\citep{chen2025gmt} adopts a mixture-of-experts design to improve motion coverage, while UniTracker~\citep{yin2025unitracker} uses a teacher-student framework for generalizable tracking. 
SONIC~\citep{luo2025sonic} scales natural humanoid whole-body tracking to large motion corpora and provides an effective controller interface for real-world humanoid execution. 
Humanoid-GPT~\citep{qi2026humanoid} further explores scaling in whole-body control by training a GPT-style causal transformer~\citep{vaswani2017attention} on a billion-scale motion corpus, achieving strong tracking performance and zero-shot generalization to unseen motions and control tasks.
\subsection{Humanoid VLAs}
Vision-language-action models (VLAs) have recently shown strong potential for language-conditioned robot control. 
The $\pi$ series~\citep{intelligence2025pi} demonstrates generalist manipulation capabilities, while GR00T~\citep{gr00tn1_2025} extends foundation-model training to humanoid robots with large-scale real and synthetic data. 
For humanoid loco-manipulation, $\Psi$-0~\citep{wei2026psi_0} proposes a staged training strategy that combines egocentric human video pretraining with real-world humanoid post-training, emphasizing the importance of high-quality domain-specific data. 
OpenHLM~\citep{hu2026openhlm} studies whole-body native humanoid VLAs and shows that directly exposing the full humanoid action space benefits long-horizon loco-manipulation. 
WholeBodyVLA~\citep{jiang2025wholebodyvla} further addresses large-space humanoid loco-manipulation by learning from low-cost action-free egocentric videos and introducing a loco-manipulation-oriented RL policy for accurate locomotion commands. 
While these methods advance humanoid VLAs, they primarily focus on direct observation-to-action learning or improving the execution interface, whereas future visual modeling remains less explored in humanoid whole-body action generation.
\subsection{Humanoid WAMs}
World action models couple action prediction with future visual dynamics to improve robot control. 
Recent works have begun to extend this idea from tabletop manipulation to humanoid loco-manipulation. 
DiT4DiT~\citep{ma2026dit4dit} jointly models video and actions with separate video and action DiTs, but its humanoid action interface still relies on a decoupled upper- and lower-body design. 
MotionWAM~\citep{zheng2026motionwam} further introduces a real-time humanoid WAM by conditioning action generation on intermediate denoising features from a video world model and predicting unified whole-body motion tokens. 
Different from these video-world-model-centered designs, $\omega$-0 treats future visual prediction as a reconstruction-free latent predictive objective for action learning. 
Rather than relying on a large video dynamics model or test-time video-to-action inversion, $\omega$-0 uses lightweight future observation embedding prediction as auxiliary supervision and directly denoises controller-compatible whole-body action latents for concurrent humanoid loco-manipulation.
\section{Method}
\label{method}

\begin{figure*}[t]
    \centering
    \includegraphics[width=\textwidth]{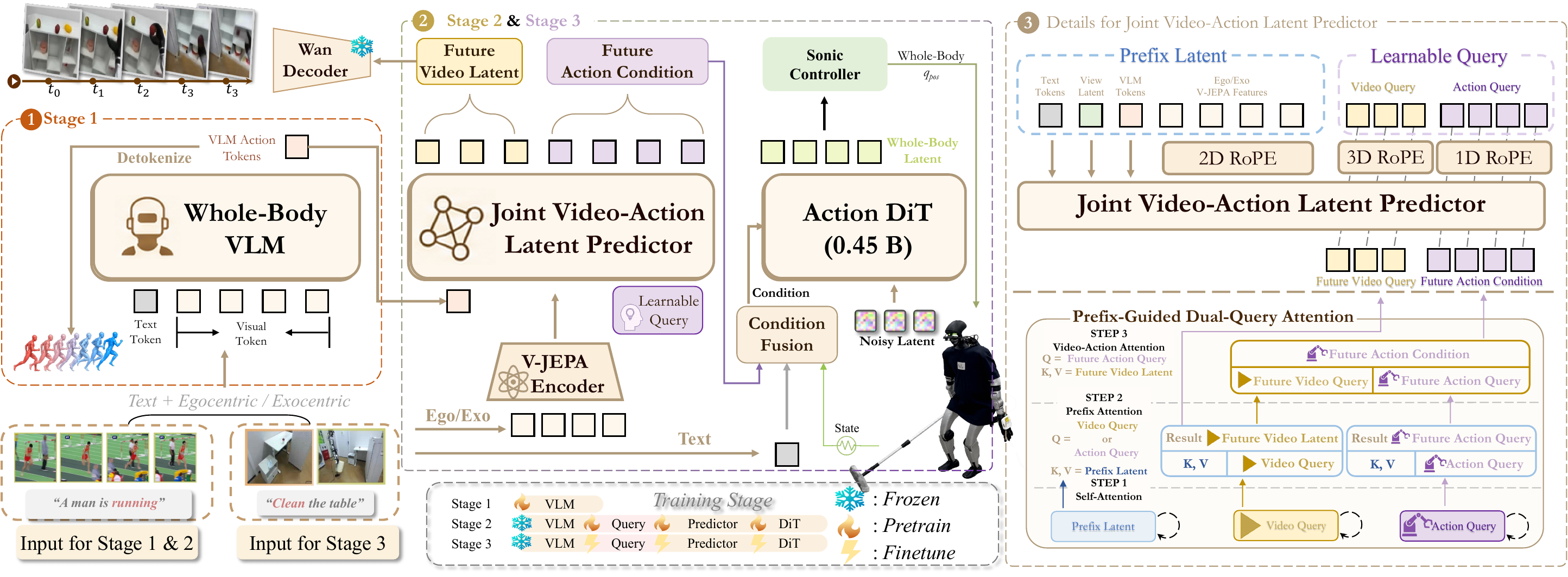}
    \caption{
    Overview of $\omega$-0.
    We first train a whole-body VLM for action-aware visual-language representation learning, and then use a joint video-action latent predictor to couple future visual latents with whole-body action generation.
    The predictor uses prefix-guided dual-query attention with token-specific positional encoding: 2D RoPE for visual prefix tokens, 3D RoPE for future video queries, and 1D RoPE for temporal action queries.
    The action DiT denoises SONIC-compatible whole-body action latents for real-world humanoid control, while the future visual branch provides lightweight latent supervision and optional visualization.
    }
    \label{framework}
\end{figure*}

This section presents the core components of $\omega$-0, as illustrated in Figure~\ref{framework}. 
We first introduce the overall architecture and problem formulation in Section~\ref{sec:method_overview}. 
Section~\ref{sec:stage1} describes whole-body action VLM pretraining, where continuous humanoid actions are converted into discrete action tokens for autoregressive visual-language learning. 
Section~\ref{sec:stage2} introduces human-to-humanoid action-latent pretraining, which aligns future visual latent prediction with diffusion-based whole-body action generation. 
Section~\ref{sec:stage3} presents real-world fine-tuning on humanoid loco-manipulation data, and Section~\ref{sec:deployment} describes the deployment pipeline with SONIC as the low-level whole-body controller.

\subsection{Overview}
\label{sec:method_overview}

We propose $\omega$-0, a whole-body world action model that maps language instructions, visual observations, and robot states to future visual latents and whole-body action latents for real-world humanoid loco-manipulation. 
Given a language instruction $\ell$, a current visual observation $\mathbf{o}_t^v$ from view $v$, and the robot proprioceptive state $\mathbf{s}_t$, the model predicts a future chunk of controller-compatible whole-body action latents $\mathbf{z}_{t:t+H}$.
The predicted latents are executed by SONIC in a receding-horizon manner on the real humanoid robot.

The model contains three main components. 
First, a whole-body action VLM maps language-conditioned visual observations into action-aware semantic features. 
To make humanoid actions compatible with autoregressive VLM training, we train a whole-body FAST tokenizer~\citep{pertsch2025fast} that discretizes continuous whole-body action trajectories into action tokens. 
Second, a joint video-action latent predictor fuses the VLM feature, T5 text feature, V-JEPA visual feature, view token, motion queries, and video queries. 
The motion queries are used for action generation, while the video queries are supervised to predict future visual latents. 
By allowing motion queries to attend to video queries, the model injects predicted visual dynamics into the action representation. 
Third, an action DiT takes the future-aware motion feature, text feature, robot state feature, and noisy action latent as input, and denoises it into a clean whole-body action latent chunk.

Unlike methods that treat visual prediction and action generation as two separate problems, $\omega$-0 uses future visual prediction as a lightweight auxiliary supervision signal for action learning. 
The model does not require a large video generator or test-time video-to-action inversion. 
Instead, it learns a shared query-based representation where future visual latents provide task-progress and scene-evolution cues for whole-body action generation. 
This enables coordinated locomotion and manipulation without explicitly decomposing tasks into separate navigation and arm-control modules.

\subsection{Stage 1: Whole-Body Action VLM Pretraining}
\label{sec:stage1}

The goal of the first stage is to endow the vision-language backbone with whole-body action semantics. 
Directly predicting continuous high-dimensional humanoid actions from visual observations is difficult for a pretrained VLM, since its output space is originally designed for discrete language tokens rather than continuous control signals. 
We therefore first construct a discrete whole-body action vocabulary, and then fine-tune a VLM to autoregressively predict action tokens from language and visual observations.

We train a whole-body FAST tokenizer on our action trajectories. 
Given a continuous whole-body action trajectory $\mathbf{a}_{t:t+H} \in \mathbb{R}^{H \times d_a}$, where $d_a$ denotes the action dimension, the tokenizer encodes the trajectory into a sequence of discrete action tokens:
\[
\mathbf{c}_{1:N} = \mathcal{E}_{\mathrm{act}}(\mathbf{a}_{t:t+H}),
\]
and a corresponding detokenizer reconstructs the original action trajectory as
\[
\hat{\mathbf{a}}_{t:t+H} = \mathcal{D}_{\mathrm{act}}(\mathbf{c}_{1:N}).
\]
The tokenizer is optimized by minimizing the reconstruction error:
\[
\mathcal{L}_{\mathrm{tok}}
=
\left\|
\hat{\mathbf{a}}_{t:t+H}
-
\mathbf{a}_{t:t+H}
\right\|_1.
\]
After training, each continuous whole-body action chunk can be represented as a compact sequence of discrete tokens, which serve as the ground-truth labels for VLM fine-tuning.

Based on the learned action tokenizer, we fine-tune Qwen3-VL-2B-Instruct~\citep{bai2025qwen3} into a whole-body action VLM. 
For each training sample, the model receives a language instruction $\ell$, a visual observation $\mathbf{o}^{v}_{t}$ from either the egocentric or exocentric view, and a learnable view token $\mathbf{e}^{v}$ that indicates the camera perspective. 
The view token is added to the visual-language input sequence to explicitly distinguish observations from different viewpoints:
\[
\mathbf{x} = [\mathbf{e}^{v}, \mathbf{o}^{v}_{t}, \ell].
\]
Conditioned on this input, the VLM is trained to autoregressively predict the discrete whole-body action tokens:
\[
p_{\theta}(\mathbf{c}_{1:N} \mid \mathbf{e}^{v}, \mathbf{o}^{v}_{t}, \ell)
=
\prod_{i=1}^{N}
p_{\theta}(c_i \mid c_{<i}, \mathbf{e}^{v}, \mathbf{o}^{v}_{t}, \ell).
\]
The training objective is the standard next-token prediction loss:
\[
\mathcal{L}_{\mathrm{vlm}}
=
-\sum_{i=1}^{N}
\log p_{\theta}(c_i \mid c_{<i}, \ell, \mathbf{o}^{v}_{t}, \mathbf{e}^{v}).
\]

After training, the whole-body action VLM learns to associate language-conditioned visual observations with discrete whole-body action tokens. 
We use its hidden representation as an action-aware semantic prior for the following stages, where it is aligned with multi-view visual features and used to condition diffusion-based whole-body latent generation.

\subsection{Stage 2: Human-to-Humanoid Action-Latent Pretraining}
\label{sec:stage2}
In the second stage, we train the model to align future visual latent prediction with whole-body action generation. 
Inspired by V-JEPA~\citep{assran2025v}, we train the model with a reconstruction-free future embedding prediction objective rather than pixel-space video generation. 
Given the current observation and action-conditioned queries, the model predicts compact future observation embeddings, providing a latent-space world-modeling signal for whole-body action generation. 
Unlike video-centered WAMs that use a video dynamics model as the main intermediate pathway for action prediction~\citep{zheng2026motionwam}, our future prediction branch is intentionally lightweight and only serves as an auxiliary predictive objective. 
The action DiT directly denoises whole-body action latents from language, visual, state, and future-aware query conditions, without test-time video-to-action inversion.

To construct supervision for this stage, we use two complementary sources. 
For future visual prediction, we extract ground-truth future visual latents using a frozen Wan encoder~\citep{wan2025}. 
For whole-body action generation, public human video-action datasets typically do not provide robot-specific action latents or proprioceptive states. 
We therefore replay their motion trajectories in simulation using SONIC. 
During replay, SONIC tracks each trajectory and produces the corresponding whole-body action latents and robot states. 
Motions that cannot be reliably executed by SONIC, such as highly dynamic or physically infeasible trajectories, are filtered out. 
This process converts generic human motion data into robot-executable supervision for diffusion-based whole-body action learning.

The resulting robot state $\mathbf{s}_t$ includes body joint positions $\mathbf{q}_{\mathrm{pos}}$, dexterous hand joint positions $\mathbf{q}_{\mathrm{hand}}$, and torso orientation from the IMU. 
Although the IMU also provides linear acceleration and angular velocity, we do not use them in our state representation. 
We only use the pelvis orientation quaternion. 
Since the same 3D rotation can be represented by two antipodal quaternions, i.e., $\mathbf{q}$ and $-\mathbf{q}$, directly using quaternions may introduce discontinuities and destabilize training. 
We therefore convert the quaternion into a continuous 6D rotation representation $\mathbf{r}_{\mathrm{torso}}^{6D}$ and define the robot state as
\[
\mathbf{s}_t =
[
\mathbf{q}_{\mathrm{pos}},
\mathbf{q}_{\mathrm{hand}},
\mathbf{r}_{\mathrm{torso}}^{6D}
].
\]

We use the frozen V-JEPA 2.1 image encoder as the visual encoder. 
Given the current visual observation $\mathbf{o}_t^v$, where $v$ denotes the egocentric or exocentric view, we extract visual features as
\[
\mathbf{f}_t^v = \mathcal{E}_{\mathrm{VJEPA}}(\mathbf{o}_t^v).
\]
The language instruction $\ell$ is encoded by a pretrained T5 encoder~\citep{raffel2020exploring} into text features $\mathbf{f}_{\ell}$. 
We further introduce a learnable view token $\mathbf{r}^v$ to indicate the camera perspective. 
Meanwhile, the whole-body action VLM pretrained in Stage 1 takes the corresponding language and visual inputs and produces an action-aware VLM feature $\mathbf{f}_{\mathrm{vlm}}$. 
We concatenate these features to form the prefix condition:
\[
\mathbf{p}
=
[
\mathbf{f}_{\mathrm{vlm}},
\mathbf{f}_{\ell},
\mathbf{r}^v,
\mathbf{f}_t^v
].
\]

To jointly predict future visual latents and motion features, we introduce two sets of learnable queries: motion queries $\mathbf{q}^{m}$ and video queries $\mathbf{q}^{v}$. 
The number of motion queries is matched to the action chunk size, so that each query corresponds to one future action step. 
The video queries correspond to future visual latent tokens and are used to predict future scene evolution in the latent space.

Since the prefix tokens, video queries, and action queries have different structures, we use token-specific RoPE~\citep{su2021roformer} in the joint predictor. 
For visual tokens in the prefix, we apply 2D RoPE according to their spatial patch coordinates. 
For future video queries, we apply 3D RoPE over temporal and spatial coordinates. 
For action queries, we apply 1D temporal RoPE along the action horizon. 
Text tokens, view tokens, and VLM summary tokens are treated as non-spatial condition tokens and are not assigned spatial RoPE. 
For an attention layer with query, key, and value $(\mathbf{Q},\mathbf{K},\mathbf{V})$, we write RoPE-based attention as
\[
\mathrm{Attn}_{\mathcal{R}}(\mathbf{Q}, \mathbf{K}, \mathbf{V})
=
\mathrm{softmax}
\left(
\frac{
\mathcal{R}(\mathbf{Q})
\mathcal{R}(\mathbf{K})^{\top}
}{\sqrt{d}}
\right)
\mathbf{V},
\]
where $\mathcal{R}$ denotes the corresponding 2D, 3D, or 1D rotary position encoding.

The prefix condition, motion queries, and video queries are first processed by separate self-attention blocks,  yielding $\tilde{\mathbf{p}}$, $\tilde{\mathbf{q}}^{m}$, and $\tilde{\mathbf{q}}^{v}$.
Then, both motion and video queries attend to the prefix condition through cross-attention:
\[
\bar{\mathbf{q}}^{m}
=
\mathrm{CrossAttn}_{m}(\tilde{\mathbf{q}}^{m}, \tilde{\mathbf{p}}),
\quad
\bar{\mathbf{q}}^{v}
=
\mathrm{CrossAttn}_{v}(\tilde{\mathbf{q}}^{v}, \tilde{\mathbf{p}}).
\]
Finally, the motion queries attend to the video queries:
\[
\mathbf{h}^{m}
=
\mathrm{CrossAttn}_{mv}(\bar{\mathbf{q}}^{m}, \bar{\mathbf{q}}^{v}),
\quad
\mathbf{h}^{v}
=
\bar{\mathbf{q}}^{v}.
\]
This interaction injects predicted visual dynamics into the motion representation, encouraging the action branch to account for the future observations induced by its own actions.

The video output $\mathbf{h}^{v}$ is supervised by the ground-truth future visual latents extracted from the frozen Wan encoder:
\[
\mathbf{y}_{t+1:t+K}^{v}
=
\mathcal{E}_{\mathrm{Wan}}(\mathbf{o}_{t+1:t+K}^{v}),
\]
with the latent prediction loss
\[
\mathcal{L}_{\mathrm{video}}
=
\left\|
\mathbf{h}^{v}
-
\mathbf{y}_{t+1:t+K}^{v}
\right\|_2^2.
\]
The predicted future visual latents are used as training-time supervision for action generation. 
When qualitative visualization is needed, we optionally decode the predicted future representation with an additional video decoder branch. 
This decoder is not used for policy inference or real-time control.

For whole-body action generation, the motion feature $\mathbf{h}^{m}$ is combined with the text feature $\mathbf{f}_{\ell}$ and the robot state feature $\mathbf{f}_{s} = \mathcal{E}_{s}(\mathbf{s}_t)$ from a state encoder. 
These features are projected into a shared hidden dimension and fused by a condition-fusion module:
\[
\mathbf{c}_{\mathrm{dit}}
=
\Phi_{\mathrm{cond}}
(
[
\mathbf{h}^{m},
\mathbf{f}_{\ell},
\mathbf{f}_{s}
]
).
\]
The fused condition is used by an action DiT~\citep{peebles2023scalable} to denoise whole-body action latents. 
Let $\mathbf{z}_0$ be the ground-truth whole-body action latent for the future action chunk obtained from SONIC replay. 
At diffusion step $\tau$, we perturb $\mathbf{z}_0$ with Gaussian noise:
\[
\mathbf{z}_{\tau}
=
\sqrt{\bar{\alpha}_{\tau}} \mathbf{z}_0
+
\sqrt{1 - \bar{\alpha}_{\tau}} \boldsymbol{\epsilon},
\quad
\boldsymbol{\epsilon} \sim \mathcal{N}(0, \mathbf{I}).
\]
The action DiT predicts the clean latent $\hat{\mathbf{z}}_0$ conditioned on the fused condition:
\[
\hat{\mathbf{z}}_0
=
\mathcal{D}_{\theta}
(
\mathbf{z}_{\tau},
\tau
\mid
\mathbf{c}_{\mathrm{dit}}
).
\]
We train the action branch with an $x_0$-prediction objective:
\[
\mathcal{L}_{\mathrm{action}}
=
\left\|
\hat{\mathbf{z}}_0
-
\mathbf{z}_0
\right\|_2^2.
\]
The overall Stage-2 objective is
\[
\mathcal{L}_{\mathrm{stage2}}
=
\mathcal{L}_{\mathrm{action}}
+
\lambda_{\mathrm{video}}
\mathcal{L}_{\mathrm{video}}.
\]
During Stage 2, the V-JEPA encoder, the Wan encoder, and the pretrained whole-body action VLM are frozen. 
We optimize the joint video-action latent predictor, the state encoder, the condition-fusion module, and the action DiT. 
At inference, we adopt a DDIM~\citep{song2020denoising} reverse denoising process to efficiently sample whole-body action latents with a small number of denoising steps.

This stage learns future-aware action queries that jointly support future visual latent prediction and diffusion-based whole-body action generation.

\subsection{Stage 3: Fine-tuning on Real-World Data}
\label{sec:stage3}

With the action-aware VLM from Stage 1 and the joint action-video latent predictor from Stage 2, we further fine-tune $\omega$-0 on real-world humanoid loco-manipulation data. 
During fine-tuning, the pretrained VLM provides action-aware semantic features from language and visual observations, while the joint video-action latent predictor produces future-aware motion features supervised by both future visual latents and action diffusion. 
We fine-tune the joint video-action latent predictor, state encoder, condition-fusion module, and action DiT on real robot trajectories, using synchronized multi-view observations, robot states, and whole-body action latents. 
The V-JEPA image encoder, the Wan encoder, and the pretrained VLM are kept frozen to preserve their visual predictive prior and action semantic prior, respectively.

The optimization mostly follows Stage 2, combining future visual latent prediction with $x_0$-prediction for whole-body action latent denoising. 
To improve temporal continuity during real-world deployment, we further introduce training-time real-time chunking (RTC)~\citep{black2025training}. 
In receding-horizon control, adjacent action chunks are not independent: the beginning of a newly predicted chunk should remain consistent with the actions generated or executed in the previous step. 
To expose the model to this setting during training, we randomly sample a prefix length $M$ and replace the first $M$ frames of the noisy action latent sequence with the corresponding clean action latents:
\[
\tilde{\mathbf{z}}_{\tau}^{1:M}
=
\mathbf{z}_{0}^{1:M},
\quad
\tilde{\mathbf{z}}_{\tau}^{M+1:H}
=
\mathbf{z}_{\tau}^{M+1:H}.
\]
The action DiT then denoises the mixed latent sequence $\tilde{\mathbf{z}}_{\tau}$, while the denoising loss is computed only on the non-prefix portion:
\[
\mathcal{L}_{\mathrm{RTC}}
=
\left\|
\hat{\mathbf{z}}_{0}^{M+1:H}
-
\mathbf{z}_{0}^{M+1:H}
\right\|_2^2.
\]
The Stage-3 objective is
\[
\mathcal{L}_{\mathrm{stage3}}
=
\mathcal{L}_{\mathrm{RTC}}
+
\lambda_{\mathrm{video}}
\mathcal{L}_{\mathrm{video}}.
\]
The clean prefix serves as a temporal anchor that guides the generation of the remaining future latents. 
This training strategy reduces discontinuities between adjacent chunks and improves the smoothness of real-world whole-body execution.

After fine-tuning, the model can be deployed in a receding-horizon manner for loco-manipulation tasks that require coordinated stepping, torso motion, arm reaching, and object interaction.

\subsection{Real-World Deployment}
\label{sec:deployment}

After training, we deploy $\omega$-0 on the real humanoid robot with SONIC as the low-level whole-body controller. 
At each control step, the robot receives a language instruction and captures real-time visual observations from the camera mounted on its head. 
Meanwhile, the robot reads its proprioceptive state, including body joint positions $\mathbf{q}_{\mathrm{pos}}$, dexterous hand joint positions $\mathbf{q}_{\mathrm{hand}}$, and torso orientation from the IMU. 
We discard the IMU linear acceleration and angular velocity, and only use the orientation quaternion. 
To avoid the discontinuity caused by the double-cover property of quaternions, we convert the quaternion into a continuous 6D rotation representation and concatenate it with the joint states to form the robot state input.

Given the language instruction, current visual observation, and robot state, $\omega$-0 predicts a chunk of whole-body action latents in a receding-horizon manner. 
To maintain temporal continuity across adjacent chunks, we use the RTC mechanism during deployment. 
Specifically, we cache the last few frames of the action chunk predicted in the previous inference step and use them as the clean prefix for the current denoising process. 
The model then generates the remaining future action latents conditioned on this prefix, making the newly predicted chunk consistent with the previously generated motion.

The generated action latent is passed to SONIC, which converts it into executable whole-body control commands for the humanoid. 
This deployment pipeline allows the robot to close the perception-action loop directly from real-time visual observations and proprioceptive feedback, enabling language-conditioned whole-body loco-manipulation on the physical platform.
\section{$\omega$-HOME Dataset}

Real-world data for humanoid loco-manipulation remains limited, especially for household tasks that require coordinated locomotion, posture adjustment, dexterous manipulation, and interaction with furniture or tools. 
To support learning whole-body world-action models in realistic domestic environments, we collect $\omega$-HOME, a multimodal household humanoid dataset containing 40.3 hours of demonstrations, 4,827 episodes, and 24 tasks recorded at 30 Hz. 
Each trajectory contains synchronized language instructions, egocentric RGB observations, exocentric RGB-D observations, robot proprioceptive states, whole-body motion references, and controller-compatible action latents.

\subsection{Tasks and Modalities}

The 24 tasks in $\omega$-HOME are organized into eight high-level capability groups, including object retrieval, surface cleaning, appliance interaction, container transfer, cloth handling, storage arrangement, mobile manipulation, and tool-based floor operation. 
These tasks involve common household objects and furniture such as tables, drawers, closets, laundry baskets, washing machines, beds, refrigerators, trash bins, and cleaning tools. 
Compared with standard tabletop manipulation datasets, $\omega$-HOME emphasizes whole-body coordination: many tasks require the robot to move its base, adjust its torso, maintain balance, and manipulate objects or tools over extended horizons.

Each episode is recorded with six synchronized modalities. 
The egocentric RGB stream is captured from the robot-mounted camera and matches the visual input used during deployment. 
The exocentric RGB-D stream is captured by a ZED depth camera and provides complementary third-person supervision for global body motion, object-scene relations, and task progress. 
The proprioceptive state records the robot configuration, including body joints, hand joints, and torso orientation, while the whole-body motion references and action latents provide supervision for learning controller-compatible behaviors. 
This multimodal design allows $\omega$-HOME to support both policy learning and future visual latent prediction.

\begin{figure*}[t]
    \centering
    \includegraphics[width=1.0\textwidth]{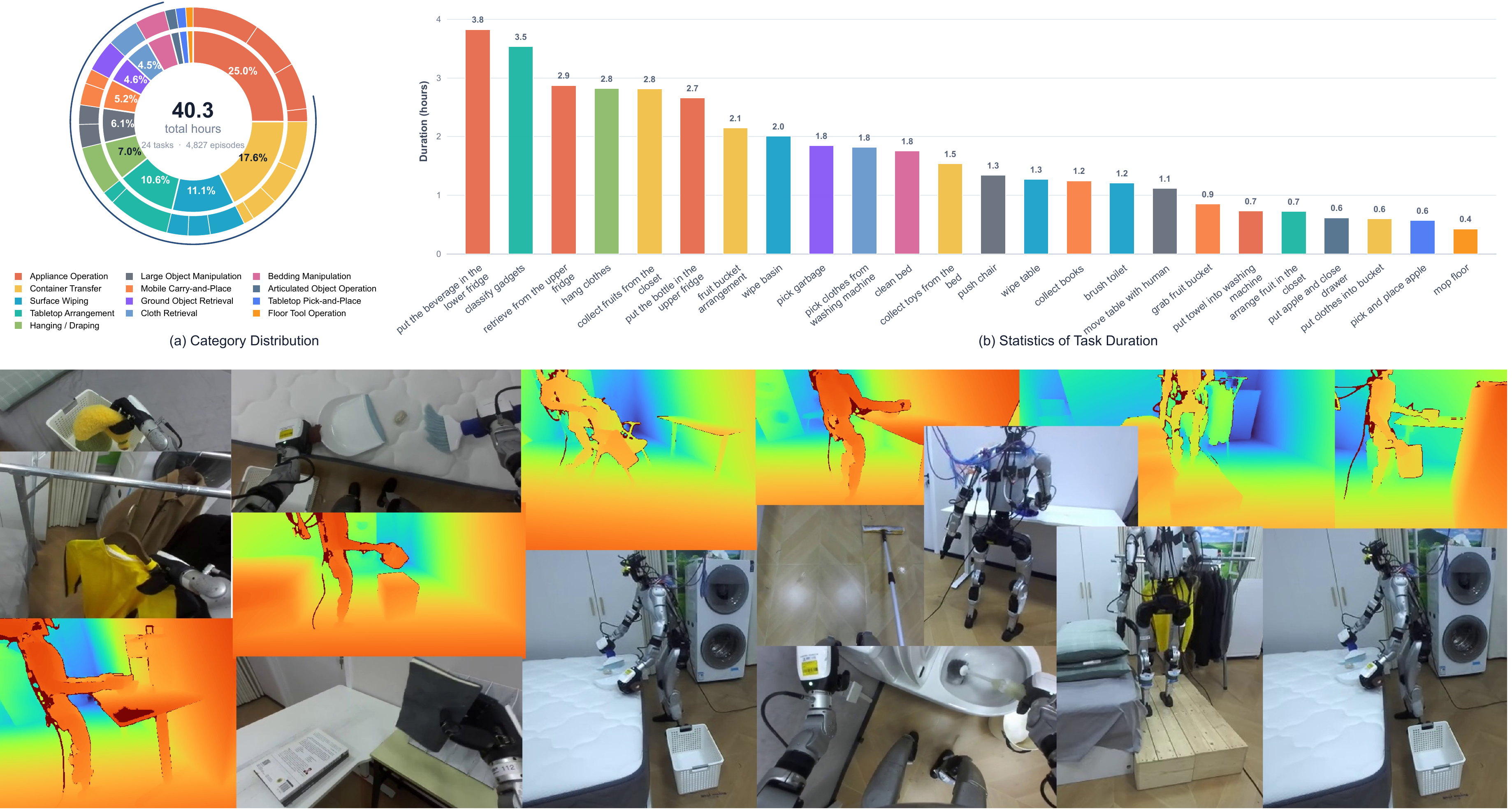}
    \caption{$\omega$-HOME dataset statistics and representative multimodal demonstrations.}
    \label{dataset_statistics}
\end{figure*}

\subsection{Dataset Statistics and Usage}
As shown in Figure~\ref{dataset_statistics}, $\omega$-HOME covers diverse household capabilities and includes both short-horizon manipulation skills and long-horizon loco-manipulation behaviors. 
The task-duration distribution shows that the dataset spans simple object interactions as well as longer activities involving appliances, cloth handling, storage, and cleaning, making it suitable for training models across different temporal horizons.

We use $\omega$-HOME for both model training and real-world evaluation. 
For downstream evaluation, we select 11 household loco-manipulation tasks and evaluate all methods under the same real-world protocol. 
During training, visual observations, robot states, and action latents supervise whole-body action generation, while future visual streams supervise future visual latent prediction. 
To evaluate the value of additional real-world humanoid data without task leakage, we exclude the 11 downstream evaluation tasks from the $\omega$-HOME pre-training pool and combine the remaining trajectories with public human demonstration data during Stage-2 pre-training. 
The pre-trained model is then fine-tuned on the downstream training split before real-world deployment. 
Empirically, incorporating this additional real-world humanoid data significantly improves real-world performance after fine-tuning, demonstrating the effectiveness of the $\omega$-HOME dataset.

\subsection{Teleoperation Details}
For teleoperation-based data collection, we utilize SONIC as the teleoperation policy. The human operator wears a Pico VR headset and two foot-mounted Pico trackers to provide head motion and lower-body motion cues. 
The operator also holds two handheld triggers, which are used to command the grasping motions of the robot hands. 
On the robot side, we use a ZED Mini camera as the onboard egocentric camera and equip the humanoid with Inspire DexHands for dexterous manipulation. 
In addition, we place a ZED depth camera in the room to synchronously capture third-person RGB observations and the corresponding depth maps. 
This setup allows us to record aligned egocentric observations, exocentric RGB-D observations, human teleoperation signals, robot states, and action trajectories for real-world whole-body loco-manipulation demonstrations.
\begin{figure}[t]
    \centering
    \includegraphics[width=0.5\textwidth]{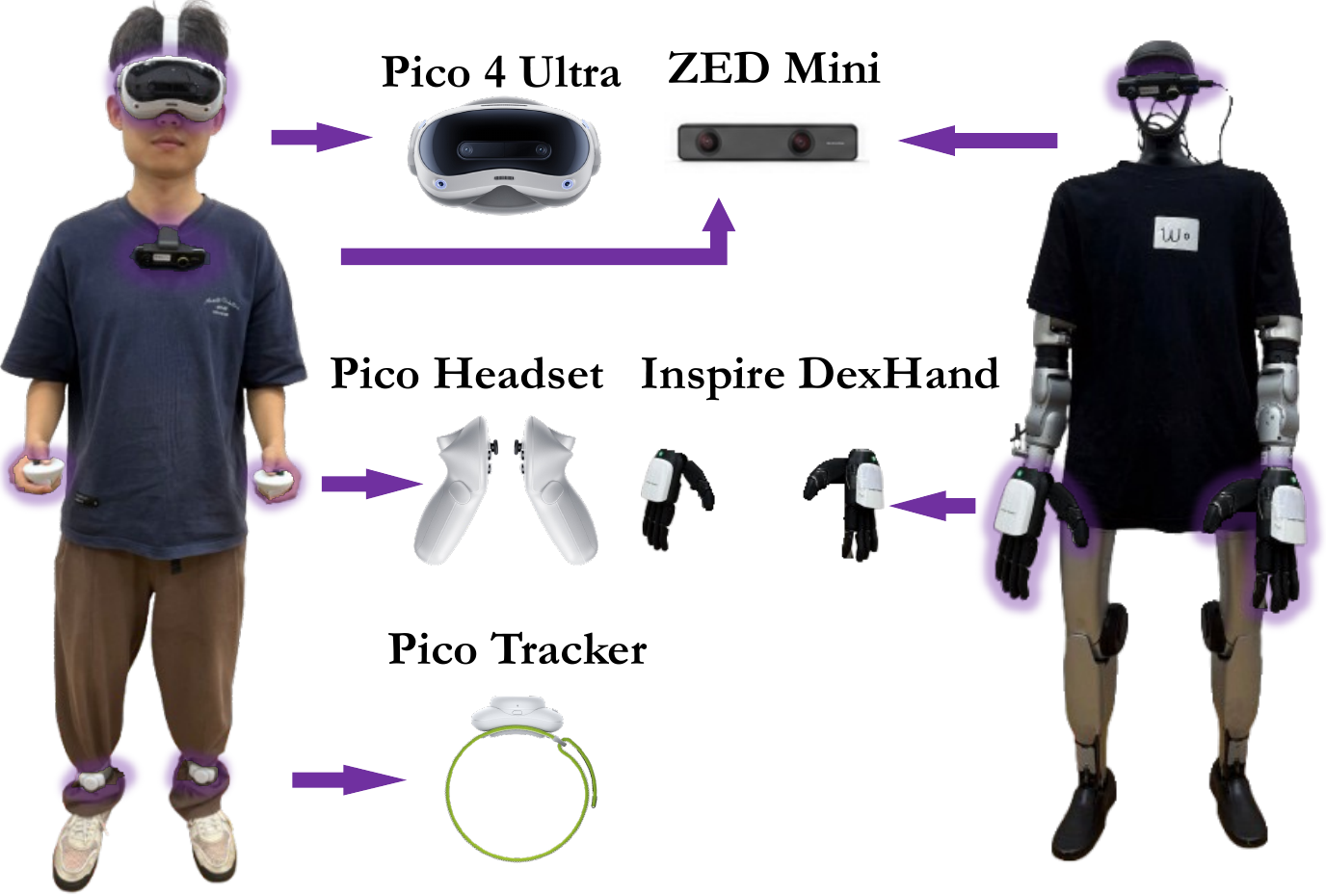}
    \caption{Real-robot teleoperation setup.
    We use a Pico 4 Ultra headset, handheld controllers, Pico trackers, and a ZED Mini camera to capture human head motion, hand commands, lower-body cues, and egocentric observations.
    The captured signals are retargeted to the humanoid with Inspire dexterous hands for whole-body loco-manipulation data collection.}
    \label{long_horizon}
\end{figure}

\section{Implementation and Experiments}
\subsection{Datasets and Data Processing}
\paragraph{\textbf{Datasets}}
Whole-body action data with paired visual observations remains limited, especially for models that must operate from both egocentric and exocentric views. To build a viewpoint-flexible whole-body action model, we combine three public datasets: ARCTIC~\citep{fan2023arctic}, Xperience-10M~\citep{xperience_10m}, and Motion-X~\citep{zhang2025motion}. ARCTIC contains both egocentric and exocentric videos, making it useful for learning view-conditioned action prediction. Xperience-10M provides large-scale egocentric multimodal data; however, not all of its sequences are relevant to humanoid loco-manipulation, so we manually filter out tasks that are unrelated to our setting or physically unsuitable for humanoid execution. Motion-X provides diverse third-person human motion videos and is used to improve exocentric action understanding.
\paragraph{\textbf{Data Processing}}
Before training, we standardize the motion representations across datasets. Since different datasets may use different body parameterizations, such as SMPL-H or SMPL-X, we convert all motion annotations into a unified SMPL representation. This preprocessing step removes dataset-specific differences in body format and allows the whole-body FAST tokenizer, VLM pretraining, and SONIC replay pipeline to operate on a consistent motion space.

In Stage 1, we merge the processed public datasets to train the whole-body action VLM. The goal of this stage is to expose the model to both egocentric and exocentric visual observations and teach it to associate language-conditioned videos with discrete whole-body action tokens. And in Stage 2, the public datasets cannot be directly used for action diffusion training because they typically provide video-action pairs but do not include robot-specific proprioceptive states or action latents that match the distribution of the SONIC controller. We therefore convert the public motions into robot-executable supervision through simulation replay. Specifically, we use SONIC to batch-replay the motion trajectories in simulation and record the corresponding robot states and whole-body action latents. During replay, trajectories that cannot be reliably executed by SONIC are discarded. This filtering is particularly important for Motion-X, which contains many highly dynamic motions such as martial-arts actions. After filtering, the replayed data provides the state-action latent pairs required for future-aware action DiT post-training.

In Stage 3, we collect real-world humanoid loco-manipulation data using SONIC-based teleoperation. Each real-world trajectory contains synchronized egocentric video, exocentric video, exocentric depth, whole-body action latents, whole-body SMPL motion, and real robot states. Unlike task-specific policy learning methods that train one model per task, we mix all real-world trajectories and fine-tune a single general model. Our training dataset covers 11 household loco-manipulation tasks, with approximately 200 demonstrations per task and 2220 trajectories in total. This mixed-task training setup encourages the model to learn a unified video-action representation across diverse behaviors, including coordinated stepping, torso motion, arm reaching, dexterous hand motion, and object interaction. After fine-tuning, the same model can be evaluated on all 11 tasks without training a separate policy for each task.
\section{Implementation Details}
We implement $\omega$-0 with Qwen3-VL-2B-Instruct as the whole-body action VLM, a frozen V-JEPA image encoder for visual latent extraction, and a pretrained T5 encoder for language features. The whole-body FAST tokenizer is trained on the unified SMPL motion representation and is used to convert continuous whole-body action trajectories into discrete action tokens for Stage-1 VLM pretraining. In Stage 2 and Stage 3, the V-JEPA encoder and the pretrained action VLM are kept frozen, while the future-aware query module, state encoder, and action DiT are optimized for future video latent prediction and whole-body action latent denoising. All three training stages are conducted on 8 NVIDIA H100 GPUs.
\subsection{Evaluation Metrics}
We evaluate all methods using three metrics: success rate, subtask score, and task progress. 
All baselines and $\omega$-0 are evaluated under the same protocol. 
For each task, we conduct 10 independent trials for every method, using a single multi-task policy trained or fine-tuned over all tasks.
\paragraph{Success Rate}
Success rate measures whether a policy can complete the entire task. 
A trial is considered successful only if all required task objectives are completed within the evaluation horizon. 
For each task, we report the success rate as $\mathrm{SR} = \frac{N_{\mathrm{success}}}{10}$, where $N_{\mathrm{success}}$ denotes the number of successful trials among 10 evaluations.
\paragraph{Score}
To capture partial task completion, we decompose each task into a set of predefined subtasks. 
Completing each subtask gives one point, and the score of a rollout is the total number of completed subtasks. 
We then report the average score over 10 trials for each task. 
The detailed subtask decomposition for each task is provided in the Appendix. 
Unlike the success rate, the score does not require the entire task to be completed. 
Moreover, an early mistake does not prevent the policy from receiving credit for later correctly executed subtasks. 
This makes the score a non-prefix measure of how many task components the policy can model within a rollout.
\paragraph{Task Progress}
Task progress measures how far a rollout proceeds along the intended task sequence before the first failure or irreversible deviation. 
For a task decomposed into $n$ ordered subtasks, if the rollout first fails at the $m$-th subtask, its task progress is defined as $\mathrm{Progress} = \frac{m}{n}$.
If the rollout completes the entire task, the progress is 1.0. 
This metric is complementary to the subtask score: while the score counts all correctly completed subtasks, task progress emphasizes continuous execution along the desired task order.
\begin{table*}[h]
\centering
\small
\setlength{\tabcolsep}{1pt} 
\begin{tabular}{clcc}
\toprule
\# & Task & Scene & Lower-body Involvement \\
\midrule
1 & Pick an apple and place into a basket& Tabletop & \textcolor{red}{\ding{55}} \\   
2 & Arrange an apple on a shelf & Shelf & \textcolor{green}{\ding{51}} \\  
3 & Pick clothes from bed and throw into basket & Bedroom & \textcolor{green}{\ding{51}} \\
4 & Move towel from basket to washing machine & Laundry & \textcolor{green}{\ding{51}} \\
5 & Wipe the table & Tabletop & \textcolor{green}{\ding{51}} \\
6 & Mop the floor & Living room & \textcolor{green}{\ding{51}} \\
7 & Pick trash from different heights into handheld bin & Household & \textcolor{green}{\ding{51}} \\
8 & Pick apple from table, throw into drawer, and close drawer with knee & Tabletop & \textcolor{green}{\ding{51}} \\
9 & Sweep trash from bed, turn around, and throw into bin & Bedroom & \textcolor{green}{\ding{51}} \\
10 & Take clothes out of the washing machine & Laundry & \textcolor{green}{\ding{51}} \\
11 & Retrieve a drink from the fridge & Fridge & \textcolor{green}{\ding{51}} \\
\bottomrule
\end{tabular}
\caption{Real-world household loco-manipulation task suite. The tasks are designed to evaluate coordinated upper- and lower-body control across tabletop, cleaning, laundry, and object-transfer scenarios.}
\label{tasks}
\end{table*}

\subsection{Baselines}
For fair comparison, all methods below are trained or fine-tuned on our real-world dataset and evaluated under the same protocol. Following our setup, each method is trained as a single multi-task model over all tasks, rather than using separate task-specific policies.
\paragraph{\textbf{ACT}}
ACT~\citep{zhao2023learning} is a transformer-based imitation learning method that predicts action chunks from visual observations and proprioceptive states. We adapt its action head to our humanoid action space and tune the chunk size for long-horizon execution. Since ACT does not use large-scale visual-language pretraining or diffusion-based action generation, it mainly serves as a classical action-chunking baseline for evaluating whether sequence prediction alone is sufficient for humanoid loco-manipulation.

\paragraph{\textbf{Diffusion Policy (DP)}}
Diffusion Policy~\citep{chi2025diffusion} models action generation as an iterative denoising process. We use a pretrained ResNet-18 as the visual encoder and train the diffusion policy to generate humanoid action chunks. During inference, the policy transforms Gaussian noise into actions through multiple denoising steps. DP provides a strong low-level imitation baseline, but its visual encoder and UNet-style action model have limited capacity for language-conditioned, long-horizon whole-body tasks.

\paragraph{\texorpdfstring{\textbf{$\pi$-0.5}}{pi-0.5}}
$\pi$-0.5~\citep{intelligence2025pi} is a generalist vision-language-action model for robot control. Since the released model is not directly configured for our humanoid action space, we expand its action head and adjust the action chunk size to match our setting. The pretrained checkpoint is used for initialization, and the modified output layers are adapted to the new action dimension. We fine-tune the resulting model on our humanoid demonstrations for comparison.

\paragraph{\textbf{InternVLA-M1}}
InternVLA-M1~\citep{chen2025internvla} is a vision-language-action framework with strong spatial grounding ability. We include it as a baseline because spatial reasoning is important for object-centric loco-manipulation. However, its pretraining is mainly focused on spatial reasoning and robotic arm manipulation, rather than humanoid whole-body control. We therefore freeze the VLM backbone and fine-tune the action head on our humanoid demonstrations.

\paragraph{\textbf{EgoVLA}}
EgoVLA~\citep{yang2025egovla} is pretrained on egocentric human manipulation videos and provides strong priors for hand-centric manipulation. Since its original action decoder focuses on wrist and hand poses, we replace the decoder with an action head compatible with our humanoid control interface and fine-tune it on our teleoperated data. EgoVLA is useful for evaluating whether egocentric manipulation pretraining transfers to humanoid loco-manipulation, especially when lower-body coordination is required.

\paragraph{\textbf{GR00T-N1.7}}
GR00T-N1.7~\citep{gr00tn1_2025} is a general-purpose robot foundation model for manipulation and loco-manipulation. We initialize from the released pretrained checkpoint and fine-tune it on our teleoperated humanoid data following the official training recipe when applicable. Since its public inference pipeline does not provide the same RTC mechanism used in our method, we evaluate it with standard sequential action-chunk inference.

\paragraph{\texorpdfstring{\textbf{$\psi$-0}}{psi-0}}
$\psi$-0~\citep{wei2026psi_0} is an arm-centric humanoid VLA pipeline that uses AMO~\citep{li2025amo} as the low-level controller for lower-body locomotion. Its high-level policy mainly predicts upper-body or arm-hand actions, while AMO handles the lower-body motion required for execution. This makes $\psi$-0 a relevant baseline for humanoid loco-manipulation, but its action generation remains structurally decoupled from whole-body control. In contrast, our model is a whole-body world action model, which predicts SONIC-compatible whole-body action latents and jointly aligns action generation with future video latent prediction.

\paragraph{\textbf{Fast-WAM}}
Fast-WAM~\citep{yuan2026fast} is a world action model that keeps video prediction as a training-time objective but removes explicit future video generation during inference. It shows that video co-training can improve action representations without test-time imagination. However, Fast-WAM is mainly evaluated on manipulation benchmarks, as well as a real-world towel-folding task, rather than whole-body humanoid loco-manipulation. We include it to compare against a WAM design that uses world modeling for representation learning but does not address full-body humanoid control.

\paragraph{\textbf{DiT4DiT}}
DiT4DiT~\citep{ma2026dit4dit} couples a video DiT and an action DiT, using intermediate denoising features from video generation to condition action prediction. It is evaluated on real-world Unitree G1 tasks. However, its action representation is still not a unified whole-body latent interface for concurrent locomotion and manipulation. We include DiT4DiT as a strong video-generation-based WAM baseline to examine how well such a design performs on tasks that require manipulation while moving, where whole-body coordination becomes critical.
\subsection{Task Suite}
We evaluate $\omega$-0 on 11 real-world household tasks that cover both tabletop manipulation and loco-manipulation. Unlike standard manipulation benchmarks that mainly focus on arm-centric object interaction, our task suite is designed to stress the coordination between the upper and lower body. Several tasks require the robot to move its base or adjust its whole-body posture while simultaneously reaching, grasping, wiping, carrying, or interacting with objects. Therefore, successful execution depends not only on accurate hand motion but also on stable stepping, torso adjustment, balance maintenance, and continuous perception-action feedback.

All tasks are designed around domestic scenarios, reflecting common household activities while emphasizing behaviors that are less represented in existing methods. In particular, we include tasks that require contact-rich interaction and coordinated locomotion-manipulation rather than isolated tabletop actions. The full task list, including task descriptions, required body coordination, and evaluation criteria, is provided in Table \ref{tasks}.

\begin{table*}[t] \centering \small \renewcommand{\arraystretch}{1.15} \begin{tabular*}{\textwidth}
{@{\extracolsep{\fill}}lccc} 
\toprule \textbf{Method} & 
\textbf{Success Rate (\%) $\uparrow$} & 
\textbf{Score (Maximum 41) $\uparrow$} & 
\textbf{Task Progress (\%) $\uparrow$} \\ 
\midrule 
\rowcolor{gray!10}
\multicolumn{4}{l}{\textit{Classical imitation learning}} \\ 
ACT & 8.2 & 10.6 & 32.4 \\ 
Diffusion Policy & 15.5 & 14.8 & 40.6 \\ 
\midrule 
\rowcolor{gray!10}
\multicolumn{4}{l}{\textit{Vision-language-action model}} \\ 
$\pi$-0.5 & 27.3 & 20.9 & 52.8 \\ InternVLA-M1 & 31.8 & 21.8 & 55.6 \\ EgoVLA & 25.5 & 18.6 & 49.1 \\ 
GR00T-N1.7 & 22.7 & 19.7 & 49.8 \\ \midrule 
\rowcolor{gray!10}
\multicolumn{4}{l}{\textit{Humanoid and world-action model}} \\ 
$\psi$-0 & \cellcolor{ntuBaseBest}44.5 & \cellcolor{ntuBaseBest}23.6 & 59.6 \\ 
Fast-WAM & 37.1 & 22.3 & 57.8 \\ 
DiT4DiT & 43.6 & 23.1 & \cellcolor{ntuBaseBest}61.0 \\ 
\midrule 
$\omega\text{-}0_{\text{Ego}}$ & \second{79.1} & \second{35.8} & \second{88.7} \\ $\omega\text{-}0_{\text{Omni}}$ & \best{81.8} & \best{36.7} & \best{90.3} \\
\bottomrule 
\end{tabular*} 
\caption{ Overall real-world evaluation across 11 household loco-manipulation tasks. All methods are trained on the real-world dataset and evaluated under the same protocol. Success rate measures the fraction of successful trials. Score measures the average number of completed subtasks, with a maximum total score of 41 over the full task suite. Task progress measures the normalized progress before the first unrecoverable failure. } \label{main_results} 
\end{table*}

\begin{figure*}[t]
    \centering
    \includegraphics[width=0.9\linewidth]{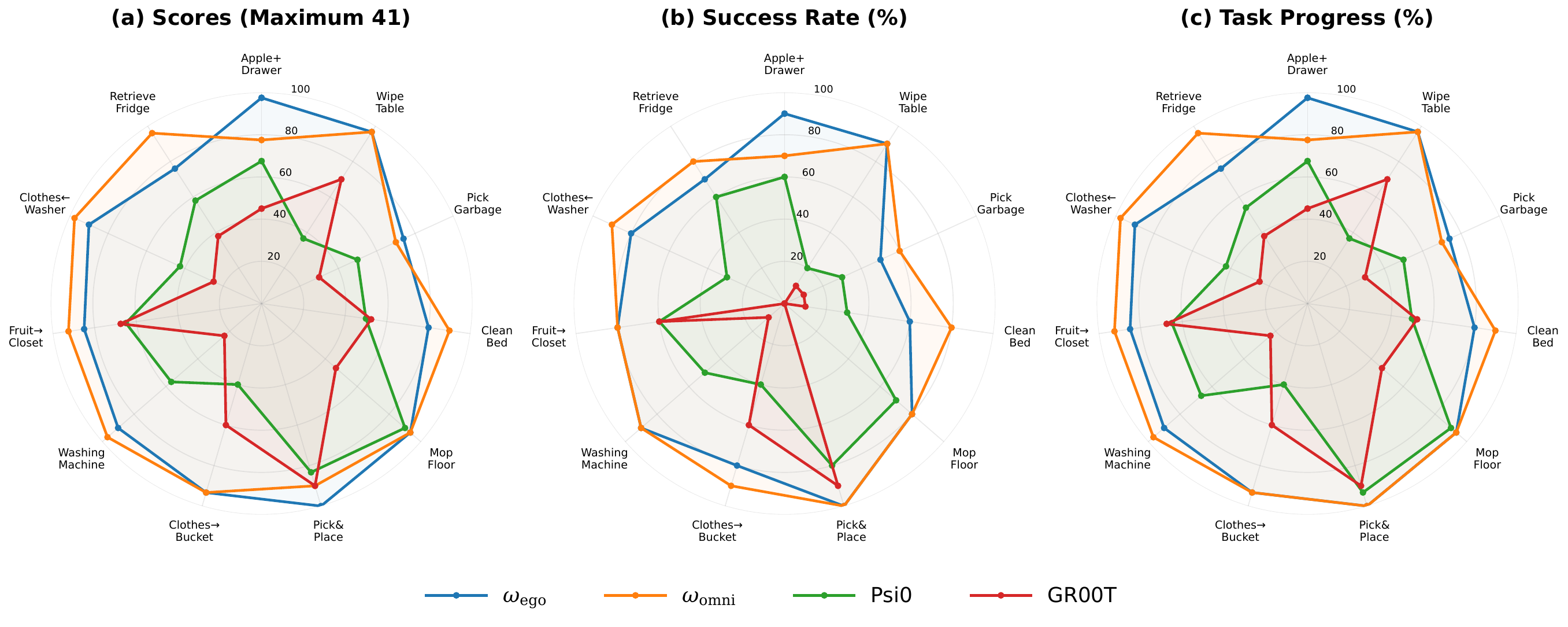}
    \caption{Real-world task setup.
    We evaluate $\omega$-0 on diverse long-horizon household loco-manipulation tasks involving whole-body motion, locomotion, tool use, articulated-object interaction, and dexterous manipulation.
    The task instruction is overlaid on each rollout sequence, and key sub-task progress is denoted with orange markers for better visualization.
    All rollouts are autonomously executed by the learned policy on the real humanoid.}
    \label{radar}
\end{figure*}

\begin{table}[t]
\centering
\small
\renewcommand{\arraystretch}{1.12}
\setlength{\tabcolsep}{3.5pt}
\begin{tabular}{lcccc}
\toprule
\textbf{Variant}
& \textbf{w/ $\omega$-HOME}
& \textbf{SR (\%) $\uparrow$}
& \textbf{Score $\uparrow$}
& \textbf{Progress (\%) $\uparrow$} \\
\midrule
$\omega\text{-}0_{\text{Ego}}$
& \rxmark
& 79.1
& 35.8
& 88.7 \\

$\omega\text{-}0_{\text{Ego}}$
& \gcmark
& \cellcolor{ntuBaseBest}80.4
& \cellcolor{ntuSecond}36.9
& \cellcolor{ntuBaseBest}89.7 \\

$\omega\text{-}0_{\text{Omni}}$
& \rxmark
& \cellcolor{ntuSecond}81.8
& \cellcolor{ntuBaseBest}36.7
& \cellcolor{ntuSecond}90.3 \\

$\omega\text{-}0_{\text{Omni}}$
& \gcmark
& \cellcolor{ntuBest}82.4
& \cellcolor{ntuBest}37.5
& \cellcolor{ntuBest}91.2 \\
\bottomrule
\end{tabular}
\caption{
Effect of using $\omega$-HOME as additional real-world humanoid pre-training data.
}
\label{data_ablation}
\end{table}

\begin{table*}[t]
\centering
\small
\renewcommand{\arraystretch}{1.12}
\setlength{\tabcolsep}{4.2pt}
\begin{tabular*}{\textwidth}{@{\extracolsep{\fill}}lcccccccc}
\toprule
\textbf{Variant}
& \textbf{State}
& \textbf{VLM Prefix}
& \textbf{Video Query}
& \textbf{RTC}
& \textbf{Image Encoder}
& \textbf{SR (\%) $\uparrow$}
& \textbf{Score $\uparrow$}
& \textbf{Progress (\%) $\uparrow$} \\
\midrule

w/o Robot state
& \rxmark
& \gcmark
& \gcmark
& \gcmark
& V-JEPA
& 60.9
& 29.8
& 75.6 \\

w/o VLM prefix
& \gcmark
& \rxmark
& \gcmark
& \gcmark
& V-JEPA
& 66.4
& 31.7
& 79.8 \\

w/o Video query
& \gcmark
& \gcmark
& \rxmark
& \gcmark
& V-JEPA
& 64.5
& 30.6
& 77.9 \\

w/o RTC
& \gcmark
& \gcmark
& \gcmark
& \rxmark
& V-JEPA
& \cellcolor{ntuBaseBest}71.8
& \cellcolor{ntuBaseBest}33.4
& \cellcolor{ntuBaseBest}84.1 \\

Wan as encoder
& \gcmark
& \gcmark
& \gcmark
& \gcmark
& Wan
& 63.6
& 30.9
& 77.3 \\

\midrule

Full $\omega\text{-}0_{\text{Ego}}$
& \gcmark
& \gcmark
& \gcmark
& \gcmark
& V-JEPA
& \second{79.1}
& \second{35.8}
& \second{88.7} \\

Full $\omega\text{-}0_{\text{Omni}}$
& \gcmark
& \gcmark
& \gcmark
& \gcmark
& V-JEPA
& \cellcolor{ntuBest}\textbf{81.8}
& \cellcolor{ntuBest}\textbf{36.7}
& \cellcolor{ntuBest}\textbf{90.3} \\

\bottomrule
\end{tabular*}
\caption{
Ablation studies of $\omega$-0 on real-world household loco-manipulation tasks.
We evaluate the effect of robot state conditioning, the Whole-Body VLM prefix, future visual latent prediction through video queries, training-time RTC, and the current-image encoder.
All variants are trained and evaluated under the same protocol.
}
\label{tab:ablation}
\end{table*}

\subsection{Main Experiment}
We evaluate $\omega$-0 and all baselines on our real-world household loco-manipulation task suite. 
The suite contains 11 tasks that require different degrees of whole-body coordination, including tabletop manipulation, object transfer, cleaning, laundry handling, and mobile manipulation. 
For fair comparison, all methods are trained or fine-tuned on our real-world dataset and evaluated under the same protocol. 
Each method is trained as a single multi-task model over all tasks, rather than using separate task-specific policies. 
For each task and each method, we conduct 10 independent trials and report success rate, subtask score, and task progress.

As shown in Table~\ref{main_results} and Figure~\ref{radar}, $\omega$-0 consistently outperforms all baselines across the main evaluation metrics. 
Classical imitation learning methods such as ACT and Diffusion Policy can generate short-horizon action chunks, but they struggle with long-horizon humanoid loco-manipulation, where the robot must coordinate locomotion, torso motion, arm motion, and dexterous manipulation. 
VLA-based baselines benefit from pretrained visual-language representations, but their action interfaces are not originally designed for unified whole-body humanoid control. 
WAM-based baselines improve action learning by incorporating future visual modeling, but they are either primarily developed for arm-centric manipulation or rely on video-generation designs that do not directly provide a controller-compatible whole-body latent interface. 
In contrast, $\omega$-0 jointly learns future visual latents and SONIC-compatible whole-body action latents, allowing it to generate more coherent whole-body behaviors. 
This leads to higher task success, better subtask completion, and larger task progress before failure.

We further compare two variants of our model to study the effect of visual viewpoint. 
The first variant, denoted as $\omega\text{-}0_{\text{Ego}}$, is trained and evaluated using only egocentric visual observations. 
The second variant, denoted as $\omega\text{-}0_{\text{Omni}}$, uses our unified visual interface and supports both egocentric and exocentric inputs. 
For $\omega\text{-}0_{\text{Omni}}$, we use exocentric observations for locomotion-heavy tasks where third-person views provide more informative motion cues, including \textit{Pick clothes from bed and throw into basket}, \textit{Move towel from basket to washing machine}, \textit{Mop the floor}, \textit{Pick trash from different heights into handheld bin}, and \textit{Pick apple from table, throw into drawer, and close drawer with knee}. 
For these tasks, the model is conditioned on exocentric images and supervised with the corresponding exocentric future visual latents. 
The remaining tasks use egocentric observations.

Table~\ref{main_results} shows that $\omega\text{-}0_{\text{Omni}}$ achieves better overall performance than $\omega\text{-}0_{\text{Ego}}$. 
This improvement is especially visible on tasks that require substantial locomotion or body repositioning. 
Egocentric observations are well aligned with execution-time perception, but they often provide a limited view of the robot's global body motion and displacement in the scene. 
For locomotion-heavy tasks, such as mopping the floor or transferring objects across locations, the first-person view may not clearly reveal the robot's stepping pattern, torso adjustment, or full-body interaction with the environment. 
In contrast, exocentric observations provide a more complete description of whole-body movement and object-scene relationships. 
By allowing the model to use exocentric inputs and exocentric future visual latent supervision for these tasks, $\omega\text{-}0_{\text{Omni}}$ learns more accurate visual-action correspondences and produces more reliable whole-body behaviors. 
These results demonstrate the effectiveness of our omni-view training design and support our claim that flexible viewpoint conditioning is beneficial for real-world humanoid loco-manipulation.

In addition, we evaluate the effectiveness of the $\omega$-HOME dataset as a source of real-world humanoid pre-training data. 
To avoid overlap between pre-training and downstream evaluation, we exclude the 11 downstream tasks used for fine-tuning from the $\omega$-HOME pre-training pool. 
The remaining $\omega$-HOME trajectories are combined with public human demonstration data and used in Stage 2 pre-training. 
After pre-training, the model is further fine-tuned on the 11 downstream household loco-manipulation tasks and deployed on the real humanoid. 
As shown in Table~\ref{data_ablation}, incorporating non-overlapping $\omega$-HOME data during pre-training leads to consistently better real-world performance after fine-tuning. 
This result demonstrates that $\omega$-HOME provides useful real-world whole-body visual-action priors beyond the downstream task demonstrations themselves, highlighting the necessity and value of our dataset.

\begin{figure*}[t]
    \centering
    \includegraphics[width=0.9\linewidth]{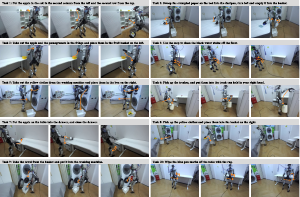}
    \caption{Real-world task setup.
    We evaluate $\omega$-0 on diverse long-horizon household loco-manipulation tasks involving whole-body motion, locomotion, tool use, articulated-object interaction, and dexterous manipulation.
    The task instruction is overlaid on each rollout sequence, and key sub-task progress is denoted with orange markers for better visualization.
    All rollouts are autonomously executed by the learned policy on the real humanoid.}
    \label{real_world}
\end{figure*}

\begin{figure*}[t]
    \centering
    \includegraphics[width=0.9\linewidth]{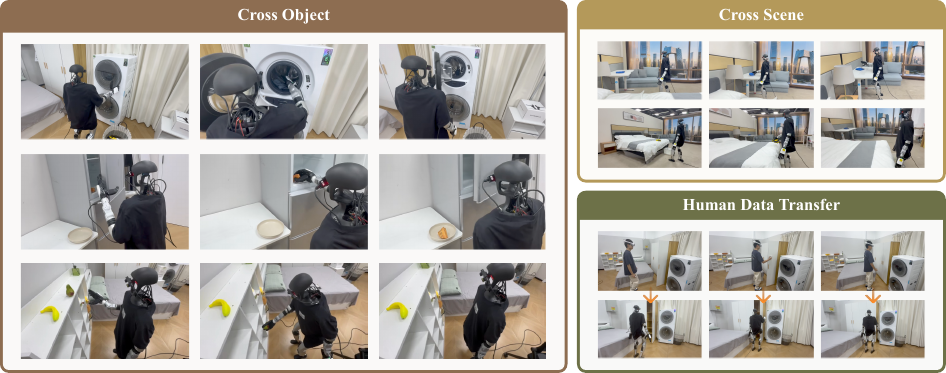}
    \caption{Generalization and human-data transfer.We evaluate $\omega$-0 on out-of-distribution objects and scenes, including novel object appearances, layouts, and household environments.
    We further fine-tune the model with human demonstration data and deploy it on the real humanoid, where it successfully completes the target tasks.}
    \label{generalization}
\end{figure*}

\begin{figure}[t]
    \centering
    \includegraphics[width=0.9\linewidth]{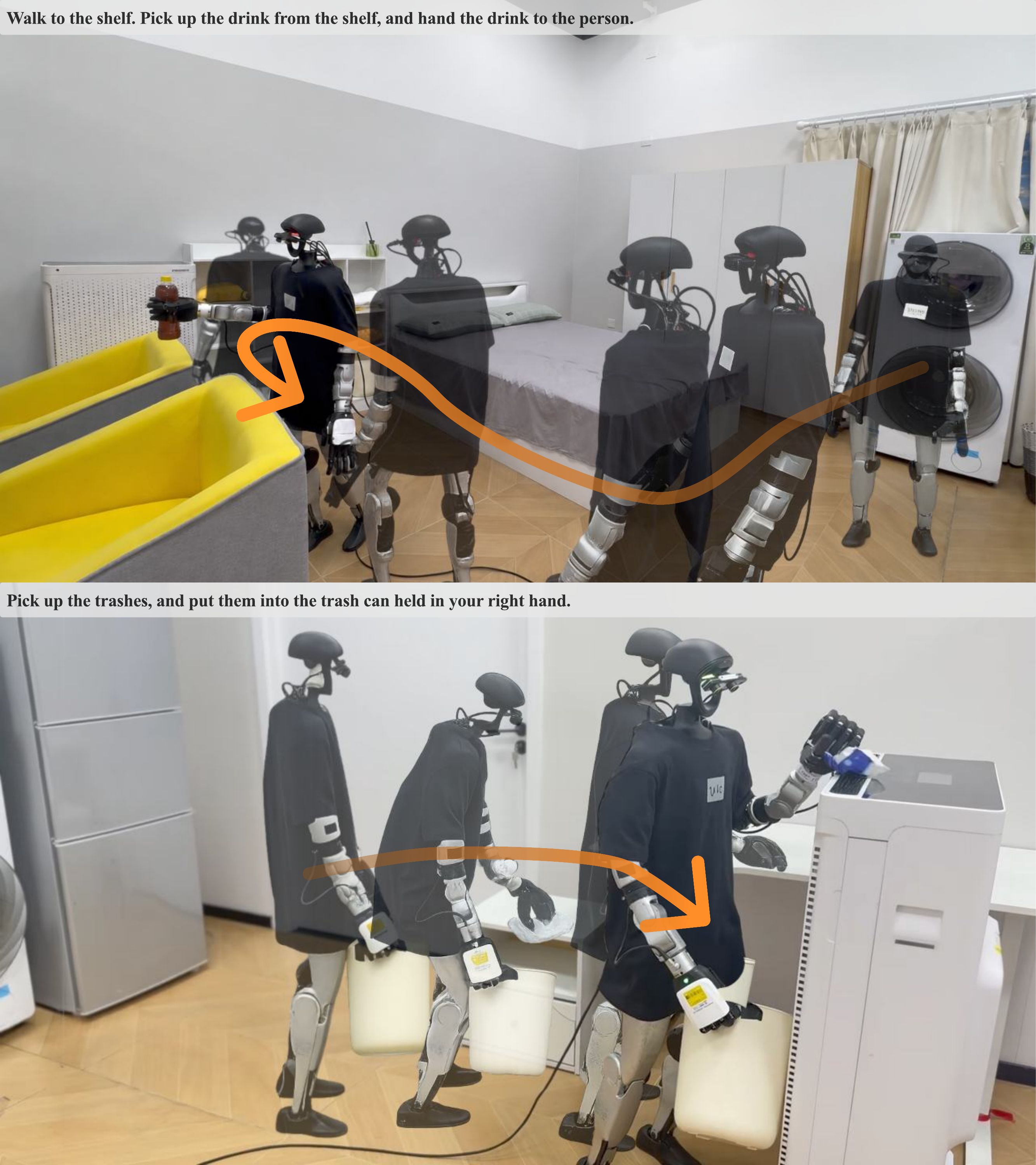}
    \caption{Real-world long-horizon tasks.}
    \label{long_horizon}
\end{figure}
\subsection{Ablation Studies}

Beyond the main experiment, we conduct a set of controlled ablations to better understand the effects of state conditioning, action-semantic prefixes, future visual latent prediction, receding-horizon chunking, and current-image encoding in $\omega$-0. 
All variants are trained and evaluated under the same protocol as the full model. 
The results are shown in Table~\ref{tab:ablation}.

\paragraph{Robot State}
We first study the effect of using the robot's current proprioceptive state. 
In real-world humanoid deployment, the same visual observation and language instruction may correspond to different actions depending on the robot's current pose, heading direction, balance condition, and hand configuration. 
Therefore, we compare the full model with a variant that removes state conditioning from the action generation module. 
Without robot state input, the policy can still rely on visual and language features, but it has less direct information about the robot's actual physical configuration. 
This particularly affects tasks involving turning, stepping, bending, or maintaining contact while moving. 
As shown in Table~\ref{tab:ablation}, removing state conditioning reduces success rate and task progress, suggesting that proprioceptive information is important for grounding the generated action latents in the current robot state.

\paragraph{Whole-Body VLM Prefix}
We then analyze the role of the whole-body VLM feature used as a prefix condition in the joint video-action latent predictor. 
This feature is produced by a VLM trained with discrete whole-body action tokens, and therefore contains task-conditioned action semantics derived from both the current visual observation and the language instruction. 
To test whether this prefix provides useful information beyond the low-level visual tokens and text features, we remove the VLM prefix while keeping the rest of the architecture unchanged. 
The resulting model shows weaker task-conditioned action generation and less accurate future visual latent prediction. 
This indicates that the VLM prefix acts as a high-level action-semantic condition, helping the predictor form more consistent future action and video latent representations.

\paragraph{Future Visual Latent Prediction}
We next ablate the future visual latent prediction branch to examine whether the world-action modeling objective is necessary for real-world control. 
In this variant, we remove the video queries, the future visual latent prediction loss, and the cross-attention from motion queries to video queries. 
The model therefore becomes a direct language-vision-state-to-action diffusion policy without latent future prediction. 
As shown in Table~\ref{tab:ablation}, removing video queries causes a clear drop in success rate, score, and task progress. 
This suggests that future visual latent prediction provides useful task-progress and scene-evolution cues for action generation, especially in long-horizon tasks that require coordinated locomotion, reaching, and contact-rich manipulation. 
The performance drop verifies that $\omega$-0 benefits not only from a stronger action decoder, but also from learning a latent predictive world-action representation.

\paragraph{Training-time RTC}
We also evaluate the effect of RTC training during action-latent prediction. 
Following prior work~\citep{wei2026psi_0}, during training, we randomly sample a prefix length between 0 and 8 and provide the corresponding ground-truth action latents as temporal context. 
The model is then trained to predict the remaining action latents. 
This procedure exposes the model to different rollout histories and better matches the receding-horizon nature of real-world deployment. 
Without RTC training, the model can still generate plausible actions, but the executed trajectories tend to contain more hesitation, discontinuity, and corrective motions. 
The ablation results show that RTC improves real-world task execution, especially in success rate and task progress.

\paragraph{Current-Image Encoder}
Finally, we compare alternative encoders for representing the current visual observation. 
One possible choice is to use the Wan encoder for both the current image and the future video latent supervision, which keeps the input and output visual latents in the same latent space. 
In offline prediction, this variant can produce more accurate future video latents. 
However, the Wan encoder is designed to encode temporally continuous video inputs and benefits from motion dynamics across frames. 
In real-world deployment, the policy receives only a single current image at each decision step. 
When this single-frame input is encoded by the Wan encoder, the visual representation tends to be less informative for action generation, and the robot often produces overly static or hesitant behavior.

In contrast, the final model uses a frozen V-JEPA encoder for the current image. 
Although this introduces a latent-space gap between the current visual tokens and the future video latent supervision, V-JEPA provides stronger single-frame semantic features for conditioning action generation. 
Empirically, this choice improves real-world task execution even when the Wan-based variant achieves better future latent reconstruction offline. 
This comparison suggests that accurate future latent reconstruction alone is not sufficient for robot control; the current-image representation must also provide action-relevant information for real-time decision making.

\paragraph{Discussion}
Overall, these ablations clarify the impact of several alternatives in our modeling pipeline. 
State conditioning improves grounding in the robot's current physical configuration; the whole-body VLM prefix provides task-conditioned action semantics; future visual latent prediction introduces a latent-space world-modeling signal that improves action generation; RTC training reduces deployment-time discontinuities under receding-horizon execution; and V-JEPA offers a more effective single-frame representation for action generation than using the video-oriented Wan encoder as the current-image encoder. 
Together, these results suggest that successful whole-body loco-manipulation depends not only on denoising action latents, but also on learning action-relevant joint embeddings that capture future task progress and scene evolution.

\subsection{Real-World Evaluation}
Figure~\ref{real_world} shows representative rollouts of $\omega$-0 on the 11 real-world loco-manipulation tasks. 
The model executes long-horizon behaviors involving whole-body posture adjustment, object interaction, and dexterous manipulation.
In particular, $\omega$-0 performs strongly on long-horizon tasks that require extended spatial movement and sustained whole-body coordination, such as collecting trash from different heights into a hand-held bin and walking a long distance to retrieve a drink from a cabinet before delivering it to a person working at a desk, as shown in Figure~\ref{long_horizon}.
All real-world robot demonstrations reported in this section are autonomously executed by the learned policy on the G1 humanoid. 
No human teleoperation, motion replay, or scripted intervention is used during evaluation.
\subsection{Generalization to Held-Out Task}
To further examine whether $\omega$-0 learns reusable whole-body visual-action priors, we evaluate its generalization under two settings, as shown in Figure~\ref{generalization}. 
First, we fine-tune $\omega$-0 on the full $\omega$-HOME dataset and evaluate it on out-of-distribution object and scene variations that are not included in the real-world fine-tuning distribution. 
Second, we study human-data transfer by fine-tuning $\omega$-0 with human demonstration data and deploying the resulting model on the real humanoid. 
These evaluations test whether the learned representation can generalize beyond the exact robot demonstrations and whether scalable human data can provide useful supervision for real-world humanoid loco-manipulation.

\paragraph{Cross-Object.}
In the cross-object setting, the task structure remains similar to the training tasks, but the target objects are replaced with novel instances or appearances. 
$\omega$-0 can still localize, approach, and manipulate these objects, suggesting that the model does not simply memorize object appearances from the training demonstrations.

\paragraph{Cross-Scene.}
In the cross-scene setting, we evaluate $\omega$-0 in new room layouts and household configurations. 
The model is required to adapt its whole-body motion, viewpoint-dependent perception, and object interaction strategy to the changed environment. 
Successful execution under these variations indicates that $\omega$-0 learns reusable visual-action priors for household loco-manipulation rather than only fitting a fixed scene layout.

\paragraph{Human Data Transfer.}
Finally, we fine-tune $\omega$-0 with human demonstration data and deploy it on the real humanoid. 
Although the demonstrations come from a different embodiment, the model can still complete the target tasks after being grounded into the robot action-latent space. 
This result highlights the potential of using scalable human data to further improve whole-body humanoid policies.

\paragraph{Effect of Video Latent Prediction on Generalization}
To further evaluate whether future video latent prediction improves the generalization ability of $\omega$-0, we conduct an additional ablation on held-out generalization settings. 
We compare the full model with a variant that removes the video query branch, including the future video latent prediction objective and the motion-to-video query interaction. 
All other components, including the VLM prefix, robot state conditioning, and action diffusion head, are kept unchanged.
The cross-object setting contains three tasks: picking a pear from a shelf, retrieving clothes with a different color from the washing machine, and taking a waffle from the refrigerator. 
The cross-scene setting contains two tasks: picking clothes from a bed in another room and walking to a table in a different scene to wipe it. 
The human-data transfer setting contains one task, where the robot walks to a closet door and closes it after being trained with human demonstration data. 
For each setting, we report the average success rate, task progress, and task score.

As shown in Table~\ref{tab:video_latent_generalization}, removing video latent prediction consistently degrades performance across all three generalization settings. 
This suggests that the video latent prediction branch provides more than an auxiliary reconstruction signal: it helps the model learn future-aware representations of task progress, object-scene interaction, and whole-body motion consequences. 
By jointly predicting future visual latents and future action conditions, $\omega$-0 obtains more transferable action representations for unseen objects, unseen scenes, and human-to-robot data transfer.
\begin{table}[t]
\centering
\small
\renewcommand{\arraystretch}{1.10}
\setlength{\tabcolsep}{3.2pt}
\resizebox{\columnwidth}{!}{
\begin{tabular}{lcccc}
\toprule
\textbf{Generalization Setting}
& \textbf{Video Query}
& \textbf{SR (\%) $\uparrow$}
& \textbf{Score $\uparrow$}
& \textbf{Progress (\%) $\uparrow$} \\
\midrule

\multirow{2}{*}{Cross-object}
& \rxmark & 66.7 & 7.6 & 63.3 \\
& \gcmark & 83.3 & 11.8 & 90.8 \\
\midrule

\multirow{2}{*}{Cross-scene}
& \rxmark & 15.0 & 0.5 & 15.0 \\
& \gcmark & 79.5 & 5.5 & 91.7 \\
\midrule

\multirow{2}{*}{Human Data Transfer}
& \rxmark & 20.0 & 1.2 & 20.0 \\
& \gcmark & 60.0 & 2.2 & 74.6 \\

\bottomrule
\end{tabular}
}
\caption{
Effect of future video latent prediction on generalization.
The maximum scores are 13 for cross-object generalization, 6 for cross-scene generalization, and 3 for human-data transfer.
}
\label{tab:video_latent_generalization}
\end{table}
\section{Conclusion}

In this work, we introduce $\omega$-0, a unified whole-body world action model for real-world humanoid concurrent loco-manipulation. 
Instead of alternating between locomotion and manipulation, $\omega$-0 generates coordinated whole-body behaviors in which the lower body, torso, arms, and hands act jointly to manipulate while moving. 
By coupling future visual latent prediction with diffusion-based whole-body action generation, $\omega$-0 learns action representations grounded in both current observations and expected scene evolution, and predicts whole-body action latents for autonomous household task execution. 
We further present $\omega$-HOME, a 40-hour multimodal household humanoid dataset with synchronized multi-view observations, robot states, whole-body motion references, and action latents. 
Real-world experiments and ablations demonstrate that $\omega$-0 consistently improves over representative imitation learning, VLA, humanoid, and WAM baselines, highlighting latent predictive world-action modeling as an effective framework for scalable concurrent humanoid loco-manipulation.

\bibliographystyle{assets/plainnat}
\bibliography{paper}

\clearpage
\onecolumn

\noindent{\Huge\bfseries Appendix}
\vspace{2em}

\appentry{A}{Data Format and Preprocessing}{app:data_format}
\appsubentry{A.1}{Action Latent and State Format}{app:action_state_format}
\appsubentry{A.2}{Normalization}{app:normalization}
\appsubentry{A.3}{Public Motion Data Processing}{app:public_motion_processing}

\appentry{B}{Inference Details and RTC Deployment}{app:inference_details}
\appsubentry{B.1}{Receding-Horizon Deployment}{app:receding_horizon}
\appsubentry{B.2}{RTC-Style Warm Start and Overlap Blending}{app:rtc_warm_start}

\appentry{C}{Real-World Evaluation Details}{app:evaluation_details}
\appsubentry{C.1}{Progress Annotation Protocol}{app:progress_annotation_protocol}
\appsubentry{C.2}{Per-Trial Progress Tables of $\omega\text{-}0_{\text{Ego}}$}{app:ego_per_trial}

\appentry{D}{Task Gallery}{app:task_gallery}

\appentry{E}{Per-Task Baseline Comparison}{app:per_task_baseline_comparison}
\appsubentry{E.1}{Per-Task Success Rate}{app:per_task_success_rate}
\appsubentry{E.2}{Per-Task Task Progress}{app:per_task_progress}
\appsubentry{E.3}{Per-Task Score}{app:per_task_score}
\clearpage
\appheading{app:data_format}{A \quad Data Format and Preprocessing}

\appsubheading{app:action_state_format}{A.1 \quad Action Latent and State Format}

Each trajectory in $\omega$-HOME is represented with controller-compatible action latents and robot proprioceptive states. 
The action representation has 66 dimensions. 
It consists of a 64-dimensional whole-body action latent produced by the low-level controller interface, together with two scalar hand commands for the left and right hands. 
For the hand commands, we use a continuous value in $[0,1]$, where 1 denotes a fully closed grasp and 0 denotes a fully open hand. 
Therefore, both the dataset action labels and the model outputs are 66-dimensional action latents.

The robot state contains body joint positions, dexterous hand joint positions, and torso orientation from the onboard IMU. 
Specifically, we use
\[
\mathbf{s}_t =
[
\mathbf{q}_{\mathrm{pos}},
\mathbf{q}_{\mathrm{hand}},
\mathbf{r}_{\mathrm{root}}^{6D}
],
\]
where $\mathbf{q}_{\mathrm{pos}}$ denotes the robot body joint positions, $\mathbf{q}_{\mathrm{hand}}$ denotes the hand joint positions, and $\mathbf{r}_{\mathrm{root}}^{6D}$ denotes the root orientation converted from the IMU quaternion to a continuous 6D rotation representation. 
We do not use IMU linear acceleration or angular velocity. 
The resulting state vector has 47 dimensions.

\appsubheading{app:normalization}{A.2 \quad Normalization}

We apply different normalization strategies to action latents and robot states for stable training and deployment. 
For the 64-dimensional whole-body action latent, we use mean-standard-deviation normalization:
\[
\tilde{\mathbf{z}} =
\frac{\mathbf{z} - \boldsymbol{\mu}_{z}}{\boldsymbol{\sigma}_{z}},
\]
where $\boldsymbol{\mu}_{z}$ and $\boldsymbol{\sigma}_{z}$ are computed from the training set. 
For the robot state and the two hand-command dimensions in the 66-dimensional action representation, we use min-max normalization:
\[
\tilde{\mathbf{x}} =
\frac{\mathbf{x} - \mathbf{x}_{\min}}
{\mathbf{x}_{\max} - \mathbf{x}_{\min}}.
\]
We use the same normalization statistics during training and inference. 
This normalization scheme improves training stability by standardizing the controller latent distribution while preserving the bounded semantics of the hand commands and proprioceptive state inputs.

\appsubheading{app:public_motion_processing}{A.3 \quad Public Motion Data Processing}

Public human-motion datasets use different motion representations and coordinate conventions. 
To obtain a unified representation, we convert both SMPL-X and SMPL-H annotations into the SMPL format before replaying them in simulation. 
We also canonicalize the coordinate system so that all motion sequences use a $z$-up convention, matching the robot simulation and deployment environment.

In addition, different motion sequences may start with different root yaw angles. 
Directly using these initial yaw values can introduce discontinuities when the motion is retargeted or replayed on the real humanoid, since the robot may experience an abrupt change in heading at the beginning of the trajectory. 
Therefore, we apply zero-yaw normalization to both public datasets and $\omega$-HOME. 
For each motion sequence, we subtract the first-frame root yaw from the root yaw of every frame:
\[
\psi_t^{\mathrm{norm}} = \psi_t - \psi_0,
\]
where $\psi_t$ is the root yaw angle at frame $t$ and $\psi_0$ is the root yaw angle of the first frame. 
This operation preserves the relative turning motion within the sequence while aligning all trajectories to a canonical initial heading. 
The processed motions are then used for SONIC-based simulation replay and action-latent extraction.

\FloatBarrier

\clearpage
\appheading{app:inference_details}{B \quad Inference Details and RTC Deployment}

\appsubheading{app:receding_horizon}{B.1 \quad Receding-Horizon Deployment}

During real-robot deployment, $\omega$-0 runs in a receding-horizon manner. 
A single forward pass takes approximately 0.14 seconds, corresponding to more than 7 Hz policy inference. 
At each inference step, the model receives the current egocentric image, the language instruction, and the current robot state, and predicts an action chunk of length $H=25$. 
Instead of executing the entire chunk before the next inference, we execute only the first $K=8$ actions and then acquire a new image for the next policy update. 
This receding-horizon strategy allows the robot to frequently refresh its visual feedback while still benefiting from temporally extended action prediction.

The text instruction is tokenized once at the beginning of an episode and reused across inference steps, while the visual tokens and robot state are refreshed at every policy update. 
For viewpoint conditioning, we use a binary view token to distinguish egocentric and exocentric observations. 
Although the model can also predict future visual latents, this branch is not required for real-time control; deployment only executes the predicted whole-body action latents.

\appsubheading{app:rtc_warm_start}{B.2 \quad RTC-Style Warm Start and Overlap Blending}

To improve temporal consistency across consecutive chunks, we use an RTC-style warm-starting strategy during inference. 
Let $\hat{\mathbf{a}}^{i}_{1:H}$ denote the action chunk predicted at inference step $i$. 
When generating the next chunk $\hat{\mathbf{a}}^{i+1}_{1:H}$, we reuse a short future segment from the previous prediction as the prefix initialization of the next denoising process, instead of initializing the whole action chunk from independent Gaussian noise. 
This encourages the next chunk to remain consistent with the previously predicted future trajectory, while still allowing the model to update its plan using the latest visual observation and proprioceptive state. 
In implementation, the prefix is taken from the unexecuted part of the previous chunk and inserted into the beginning of the next diffusion initialization.

We further apply overlap blending to reduce discontinuities at chunk boundaries. 
Given an overlap length $O$, the last $O$ actions of the previous chunk and the first $O$ actions of the next chunk are linearly blended:
\[
\mathbf{a}_{j}^{\mathrm{blend}}
=
(1-\alpha_j)\mathbf{a}_{j}^{\mathrm{prev}}
+
\alpha_j \mathbf{a}_{j}^{\mathrm{next}},
\qquad
\alpha_j = \frac{j+1}{O+1},
\quad j=0,\ldots,O-1.
\]
This simple blending removes high-frequency jumps caused by independently sampled chunks and produces smoother whole-body transitions during closed-loop execution. 
After stitching, the normalized action latents are denormalized and sent to the low-level whole-body controller for execution.

\begin{algorithm}[htbp]
\caption{RTC-style Receding-Horizon Inference}
\label{alg:rtc_inference}
\begin{algorithmic}[1]
\STATE Initialize previous prefix $\mathbf{p}\leftarrow \emptyset$
\STATE Tokenize language instruction once
\FOR{each policy update step $i$}
    \STATE Acquire current image $\mathbf{o}_i$ and robot state $\mathbf{s}_i$
    \STATE Initialize action chunk noise $\mathbf{z}_i \sim \mathcal{N}(0,I)$
    \IF{$\mathbf{p}$ is not empty}
        \STATE Replace the first prefix region of $\mathbf{z}_i$ with $\mathbf{p}$
    \ENDIF
    \STATE Predict action chunk $\hat{\mathbf{a}}^i_{1:H}$ with $\omega$-0
    \STATE Execute the first $K$ actions on the robot
    \STATE Store an unexecuted future segment of $\hat{\mathbf{a}}^i_{1:H}$ as the next prefix $\mathbf{p}$
    \STATE Blend overlapping actions between consecutive chunks
\ENDFOR
\end{algorithmic}
\end{algorithm}

\FloatBarrier

\clearpage
\appheading{app:evaluation_details}{C \quad Real-World Evaluation Details}

\appsubheading{app:progress_annotation_protocol}{C.1 \quad Progress Annotation Protocol}

We provide detailed per-trial progress annotations of $\omega\text{-}0_{\text{Ego}}$ in Tables~\ref{tab:ego_put_apple_drawer}--\ref{tab:ego_retrieve_fridge}. 
Each task is evaluated over 10 real-robot trials. 
For each task, we define a sequence of binary progress stages corresponding to key physical milestones, such as grasping the target object, transporting it to the goal region, completing placement, cleaning the target area, or closing an articulated object. 
An entry of 1 indicates that the corresponding stage is completed in that trial, while 0 indicates failure to complete that stage.

\appsubheading{app:ego_per_trial}{C.2 \quad Per-Trial Progress Tables of $\omega\text{-}0_{\text{Ego}}$}

\begin{table}[htbp]
\centering
\scriptsize
\renewcommand{\arraystretch}{1.08}
\setlength{\tabcolsep}{3.2pt}
\resizebox{\textwidth}{!}{
\begin{tabular}{lcccccccccc}
\toprule
\textbf{put\_apple\_and\_close\_drawer} 
& \textbf{Trial 1} & \textbf{Trial 2} & \textbf{Trial 3} & \textbf{Trial 4} & \textbf{Trial 5} 
& \textbf{Trial 6} & \textbf{Trial 7} & \textbf{Trial 8} & \textbf{Trial 9} & \textbf{Trial 10} \\
\midrule
Stage 1: Correctly grasp the apple
& 1 & 1 & 1 & 1 & 1 & 1 & 1 & 1 & 1 & 1 \\
Stage 2: Transport the apple to the target drawer region
& 1 & 1 & 1 & 1 & 1 & 1 & 1 & 1 & 1 & 1 \\
Stage 3: Place the apple completely inside the drawer
& 1 & 1 & 1 & 1 & 1 & 1 & 1 & 1 & 1 & 1 \\
Stage 4: Fully close the drawer
& 1 & 1 & 1 & 1 & 1 & 1 & 0 & 1 & 1 & 1 \\
\bottomrule
\end{tabular}
}
\caption{
Per-trial progress annotations of $\omega\text{-}0_{\text{Ego}}$ on \texttt{put\_apple\_and\_close\_drawer}.
}
\label{tab:ego_put_apple_drawer}
\end{table}

\begin{table}[htbp]
\centering
\scriptsize
\renewcommand{\arraystretch}{1.08}
\setlength{\tabcolsep}{3.2pt}
\resizebox{\textwidth}{!}{
\begin{tabular}{lcccccccccc}
\toprule
\textbf{wipe\_table} 
& \textbf{Trial 1} & \textbf{Trial 2} & \textbf{Trial 3} & \textbf{Trial 4} & \textbf{Trial 5} 
& \textbf{Trial 6} & \textbf{Trial 7} & \textbf{Trial 8} & \textbf{Trial 9} & \textbf{Trial 10} \\
\midrule
Stage 1: Correctly grasp the cloth
& 1 & 1 & 1 & 1 & 1 & 1 & 1 & 1 & 1 & 1 \\
Stage 2: Make effective wiping contact with the blue-marked region
& 1 & 1 & 1 & 1 & 1 & 1 & 1 & 1 & 1 & 1 \\
Stage 3: Remove all specified blue marks
& 1 & 1 & 0 & 1 & 1 & 1 & 1 & 1 & 1 & 1 \\
\bottomrule
\end{tabular}
}
\caption{
Per-trial progress annotations of $\omega\text{-}0_{\text{Ego}}$ on \texttt{wipe\_table}.
}
\label{tab:ego_wipe_table}
\end{table}

\begin{table}[htbp]
\centering
\scriptsize
\renewcommand{\arraystretch}{1.08}
\setlength{\tabcolsep}{3.2pt}
\resizebox{\textwidth}{!}{
\begin{tabular}{lcccccccccc}
\toprule
\textbf{pick\_garbage} 
& \textbf{Trial 1} & \textbf{Trial 2} & \textbf{Trial 3} & \textbf{Trial 4} & \textbf{Trial 5} 
& \textbf{Trial 6} & \textbf{Trial 7} & \textbf{Trial 8} & \textbf{Trial 9} & \textbf{Trial 10} \\
\midrule
Stage 1: Stably hold the trash bin with the right hand
& 1 & 1 & 1 & 1 & 1 & 1 & 1 & 1 & 1 & 1 \\
Stage 2: Place the first target object into the bin
& 0 & 0 & 0 & 1 & 0 & 0 & 1 & 1 & 1 & 1 \\
Stage 3: Place the second target object into the bin
& 1 & 1 & 0 & 1 & 1 & 1 & 1 & 1 & 1 & 1 \\
Stage 4: Place the third target object into the bin
& 0 & 1 & 1 & 1 & 0 & 1 & 1 & 1 & 1 & 1 \\
Stage 5: Keep all target objects inside the bin with no distractor object
& 0 & 0 & 0 & 1 & 0 & 0 & 1 & 1 & 1 & 1 \\
\bottomrule
\end{tabular}
}
\caption{
Per-trial progress annotations of $\omega\text{-}0_{\text{Ego}}$ on \texttt{pick\_garbage}.
}
\label{tab:ego_pick_garbage}
\end{table}

\begin{table}[htbp]
\centering
\scriptsize
\renewcommand{\arraystretch}{1.08}
\setlength{\tabcolsep}{3.2pt}
\resizebox{\textwidth}{!}{
\begin{tabular}{lcccccccccc}
\toprule
\textbf{clean\_bed} 
& \textbf{Trial 1} & \textbf{Trial 2} & \textbf{Trial 3} & \textbf{Trial 4} & \textbf{Trial 5} 
& \textbf{Trial 6} & \textbf{Trial 7} & \textbf{Trial 8} & \textbf{Trial 9} & \textbf{Trial 10} \\
\midrule
Stage 1: Correctly obtain the brush and dustpan
& 1 & 1 & 1 & 1 & 1 & 1 & 1 & 1 & 1 & 1 \\
Stage 2: Sweep all paper balls on the bed into the dustpan
& 1 & 1 & 0 & 1 & 1 & 1 & 1 & 1 & 1 & 0 \\
Stage 3: Turn toward the left basket while carrying the paper balls
& 1 & 1 & 0 & 1 & 1 & 1 & 1 & 1 & 1 & 0 \\
Stage 4: Dump all paper balls into the basket
& 1 & 1 & 0 & 0 & 1 & 1 & 1 & 1 & 0 & 0 \\
\bottomrule
\end{tabular}
}
\caption{
Per-trial progress annotations of $\omega\text{-}0_{\text{Ego}}$ on \texttt{clean\_bed}.
}
\label{tab:ego_clean_bed}
\end{table}

\begin{table}[htbp]
\centering
\scriptsize
\renewcommand{\arraystretch}{1.08}
\setlength{\tabcolsep}{3.2pt}
\resizebox{\textwidth}{!}{
\begin{tabular}{lcccccccccc}
\toprule
\textbf{mop\_floor} 
& \textbf{Trial 1} & \textbf{Trial 2} & \textbf{Trial 3} & \textbf{Trial 4} & \textbf{Trial 5} 
& \textbf{Trial 6} & \textbf{Trial 7} & \textbf{Trial 8} & \textbf{Trial 9} & \textbf{Trial 10} \\
\midrule
Stage 1: Correctly grasp the mop
& 1 & 1 & 1 & 1 & 1 & 1 & 1 & 1 & 1 & 1 \\
Stage 2: Make effective mopping contact with the target region
& 1 & 1 & 1 & 1 & 1 & 1 & 1 & 1 & 1 & 1 \\
Stage 3: Remove all specified black water stains
& 1 & 1 & 1 & 1 & 1 & 1 & 0 & 0 & 1 & 1 \\
\bottomrule
\end{tabular}
}
\caption{
Per-trial progress annotations of $\omega\text{-}0_{\text{Ego}}$ on \texttt{mop\_floor}.
}
\label{tab:ego_mop_floor}
\end{table}

\begin{table}[htbp]
\centering
\scriptsize
\renewcommand{\arraystretch}{1.08}
\setlength{\tabcolsep}{3.2pt}
\resizebox{\textwidth}{!}{
\begin{tabular}{lcccccccccc}
\toprule
\textbf{pick\_and\_place} 
& \textbf{Trial 1} & \textbf{Trial 2} & \textbf{Trial 3} & \textbf{Trial 4} & \textbf{Trial 5} 
& \textbf{Trial 6} & \textbf{Trial 7} & \textbf{Trial 8} & \textbf{Trial 9} & \textbf{Trial 10} \\
\midrule
Stage 1: Correctly grasp the apple
& 1 & 1 & 1 & 1 & 1 & 1 & 1 & 1 & 1 & 1 \\
Stage 2: Transport the apple to the fruit-basket opening
& 1 & 1 & 1 & 1 & 1 & 1 & 1 & 1 & 1 & 1 \\
Stage 3: Release the apple and keep it inside the fruit basket
& 1 & 1 & 1 & 1 & 1 & 1 & 1 & 1 & 1 & 1 \\
\bottomrule
\end{tabular}
}
\caption{
Per-trial progress annotations of $\omega\text{-}0_{\text{Ego}}$ on \texttt{pick\_and\_place}.
}
\label{tab:ego_pick_and_place}
\end{table}

\begin{table}[htbp]
\centering
\scriptsize
\renewcommand{\arraystretch}{1.08}
\setlength{\tabcolsep}{3.2pt}
\resizebox{\textwidth}{!}{
\begin{tabular}{lcccccccccc}
\toprule
\textbf{put\_clothes\_into\_bucket} 
& \textbf{Trial 1} & \textbf{Trial 2} & \textbf{Trial 3} & \textbf{Trial 4} & \textbf{Trial 5} 
& \textbf{Trial 6} & \textbf{Trial 7} & \textbf{Trial 8} & \textbf{Trial 9} & \textbf{Trial 10} \\
\midrule
Stage 1: Correctly grasp the yellow cloth on the bed
& 1 & 1 & 1 & 1 & 1 & 1 & 1 & 1 & 1 & 1 \\
Stage 2: Transport the yellow cloth above the right white basket
& 1 & 1 & 1 & 1 & 1 & 1 & 1 & 1 & 1 & 1 \\
Stage 3: Place the yellow cloth completely into the white basket
& 1 & 1 & 1 & 1 & 0 & 1 & 1 & 1 & 1 & 0 \\
\bottomrule
\end{tabular}
}
\caption{
Per-trial progress annotations of $\omega\text{-}0_{\text{Ego}}$ on \texttt{put\_clothes\_into\_bucket}.
}
\label{tab:ego_put_clothes_bucket}
\end{table}

\begin{table}[htbp]
\centering
\scriptsize
\renewcommand{\arraystretch}{1.08}
\setlength{\tabcolsep}{3.2pt}
\resizebox{\textwidth}{!}{
\begin{tabular}{lcccccccccc}
\toprule
\textbf{washing\_machine} 
& \textbf{Trial 1} & \textbf{Trial 2} & \textbf{Trial 3} & \textbf{Trial 4} & \textbf{Trial 5} 
& \textbf{Trial 6} & \textbf{Trial 7} & \textbf{Trial 8} & \textbf{Trial 9} & \textbf{Trial 10} \\
\midrule
Stage 1: Correctly pick up the towel from the basket
& 1 & 1 & 1 & 1 & 1 & 1 & 0 & 1 & 1 & 1 \\
Stage 2: Transport the towel to the washing-machine opening
& 1 & 1 & 1 & 1 & 1 & 1 & 0 & 1 & 1 & 1 \\
Stage 3: Place the towel completely inside the washing machine
& 1 & 1 & 1 & 1 & 1 & 1 & 0 & 1 & 1 & 1 \\
\bottomrule
\end{tabular}
}
\caption{
Per-trial progress annotations of $\omega\text{-}0_{\text{Ego}}$ on \texttt{washing\_machine}.
}
\label{tab:ego_washing_machine}
\end{table}

\begin{table}[htbp]
\centering
\scriptsize
\renewcommand{\arraystretch}{1.08}
\setlength{\tabcolsep}{3.2pt}
\resizebox{\textwidth}{!}{
\begin{tabular}{lcccccccccc}
\toprule
\textbf{arrange\_fruit\_in\_the\_closet} 
& \textbf{Trial 1} & \textbf{Trial 2} & \textbf{Trial 3} & \textbf{Trial 4} & \textbf{Trial 5} 
& \textbf{Trial 6} & \textbf{Trial 7} & \textbf{Trial 8} & \textbf{Trial 9} & \textbf{Trial 10} \\
\midrule
Stage 1: Correctly grasp the apple
& 1 & 1 & 1 & 1 & 0 & 1 & 1 & 1 & 1 & 1 \\
Stage 2: Transport the apple near the target closet cell
& 1 & 1 & 1 & 1 & 0 & 1 & 1 & 1 & 1 & 1 \\
Stage 3: Align the apple with the specified closet cell
& 1 & 1 & 1 & 1 & 0 & 1 & 0 & 1 & 1 & 1 \\
Stage 4: Stably place the apple into the specified cell
& 1 & 1 & 1 & 1 & 0 & 1 & 0 & 1 & 1 & 1 \\
\bottomrule
\end{tabular}
}
\caption{
Per-trial progress annotations of $\omega\text{-}0_{\text{Ego}}$ on \texttt{arrange\_fruit\_in\_the\_closet}.
}
\label{tab:ego_arrange_fruit_closet}
\end{table}

\begin{table}[htbp]
\centering
\scriptsize
\renewcommand{\arraystretch}{1.08}
\setlength{\tabcolsep}{3.2pt}
\resizebox{\textwidth}{!}{
\begin{tabular}{lcccccccccc}
\toprule
\textbf{pick\_clothes\_from\_washing\_machine} 
& \textbf{Trial 1} & \textbf{Trial 2} & \textbf{Trial 3} & \textbf{Trial 4} & \textbf{Trial 5} 
& \textbf{Trial 6} & \textbf{Trial 7} & \textbf{Trial 8} & \textbf{Trial 9} & \textbf{Trial 10} \\
\midrule
Stage 1: Open the washing-machine door
& 1 & 1 & 1 & 1 & 1 & 1 & 1 & 1 & 1 & 1 \\
Stage 2: Completely take out the yellow cloth from the washing machine
& 1 & 1 & 1 & 1 & 1 & 1 & 1 & 1 & 1 & 1 \\
Stage 3: Place the yellow cloth completely into the right box
& 1 & 1 & 0 & 1 & 1 & 1 & 1 & 0 & 1 & 1 \\
Stage 4: Fully close the washing-machine door
& 1 & 1 & 0 & 1 & 1 & 1 & 1 & 0 & 1 & 1 \\
\bottomrule
\end{tabular}
}
\caption{
Per-trial progress annotations of $\omega\text{-}0_{\text{Ego}}$ on \texttt{pick\_clothes\_from\_washing\_machine}.
}
\label{tab:ego_pick_clothes_washing_machine}
\end{table}

\begin{table}[htbp]
\centering
\scriptsize
\renewcommand{\arraystretch}{1.08}
\setlength{\tabcolsep}{3.2pt}
\resizebox{\textwidth}{!}{
\begin{tabular}{lcccccccccc}
\toprule
\textbf{retrieve\_from\_fridge} 
& \textbf{Trial 1} & \textbf{Trial 2} & \textbf{Trial 3} & \textbf{Trial 4} & \textbf{Trial 5} 
& \textbf{Trial 6} & \textbf{Trial 7} & \textbf{Trial 8} & \textbf{Trial 9} & \textbf{Trial 10} \\
\midrule
Stage 1: Open the refrigerator door
& 1 & 1 & 1 & 1 & 1 & 1 & 1 & 1 & 1 & 1 \\
Stage 2: Place the first fruit into the left fruit basket
& 1 & 1 & 1 & 1 & 0 & 1 & 0 & 1 & 0 & 1 \\
Stage 3: Place the second fruit into the left fruit basket
& 1 & 1 & 1 & 1 & 0 & 1 & 0 & 1 & 0 & 1 \\
Stage 4: Keep both fruits stably inside the fruit basket
& 1 & 1 & 1 & 1 & 0 & 1 & 0 & 1 & 0 & 1 \\
Stage 5: Fully close the refrigerator door
& 1 & 1 & 1 & 1 & 0 & 1 & 0 & 1 & 0 & 1 \\
\bottomrule
\end{tabular}
}
\caption{
Per-trial progress annotations of $\omega\text{-}0_{\text{Ego}}$ on \texttt{retrieve\_from\_fridge}.
}
\label{tab:ego_retrieve_fridge}
\end{table}

\FloatBarrier

\appheading{app:per_task_baseline_comparison}{E \quad Per-Task Baseline Comparison}

In addition to the aggregate real-world results reported in the main paper, we provide a more fine-grained per-task comparison across representative baselines. 
Specifically, we compare seven models, including $\omega\text{-}0_{\text{Ego}}$, $\omega\text{-}0_{\text{Omni}}$, $\psi$-0, GROOT N1.7, Fast-WAM, DiT4DiT, and $\pi$-0.5. 
For each method, we report per-task success rate, task progress, and average task score over the 11 real-world household loco-manipulation tasks.

These per-task results reveal several trends that are partially hidden by aggregate metrics. 
First, $\omega\text{-}0_{\text{Ego}}$ and $\omega\text{-}0_{\text{Omni}}$ consistently achieve strong performance across both short-horizon manipulation tasks and long-horizon loco-manipulation tasks. 
Second, baseline methods can perform reasonably well on simpler tasks such as pick-and-place, but their performance drops more significantly on tasks that require articulated-object interaction, sequential object collection, dual-hand coordination, or extended whole-body long-distance locomotion. 
Third, the omni-view variant further improves several challenging tasks by leveraging complementary visual observations, suggesting that multi-view supervision provides useful information for whole-body task progress and scene understanding.

\appsubheading{app:per_task_success_rate}{E.1 \quad Per-Task Success Rate}

\begin{figure}[p]
    \centering
    \includegraphics[width=1.0\textwidth]{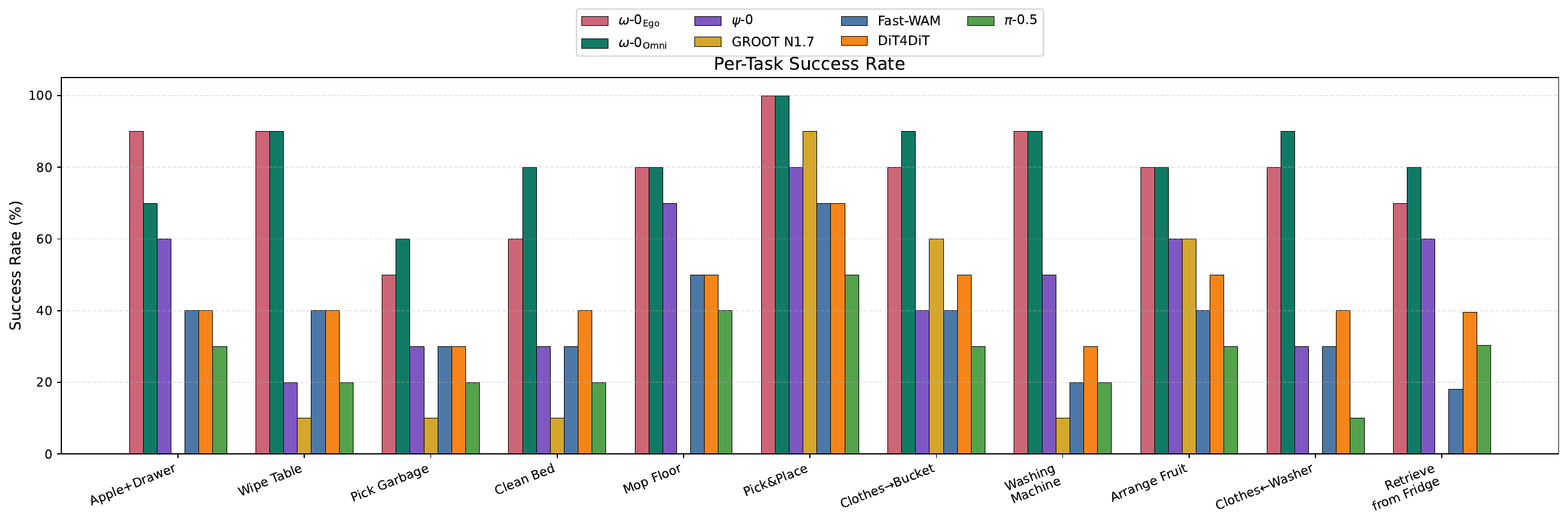}
    \caption{
    Per-task success rate comparison across seven models on 11 real-world household loco-manipulation tasks. 
    Success rate is computed as the percentage of trials in which all progress stages of a task are completed.
    }
    \label{fig:all7models_success_rate}
\end{figure}

Figure~\ref{fig:all7models_success_rate} shows the per-task success rate of all seven models. 
Success rate measures the percentage of trials in which all predefined progress stages are completed. 
The results show that $\omega\text{-}0_{\text{Ego}}$ and $\omega\text{-}0_{\text{Omni}}$ achieve higher success rates on most tasks, especially on tasks involving contact-rich cleaning, articulated-object operation, and long-horizon retrieval.

\appsubheading{app:per_task_progress}{E.2 \quad Per-Task Task Progress}

\begin{figure}[p]
    \centering
    \includegraphics[width=1.0\textwidth]{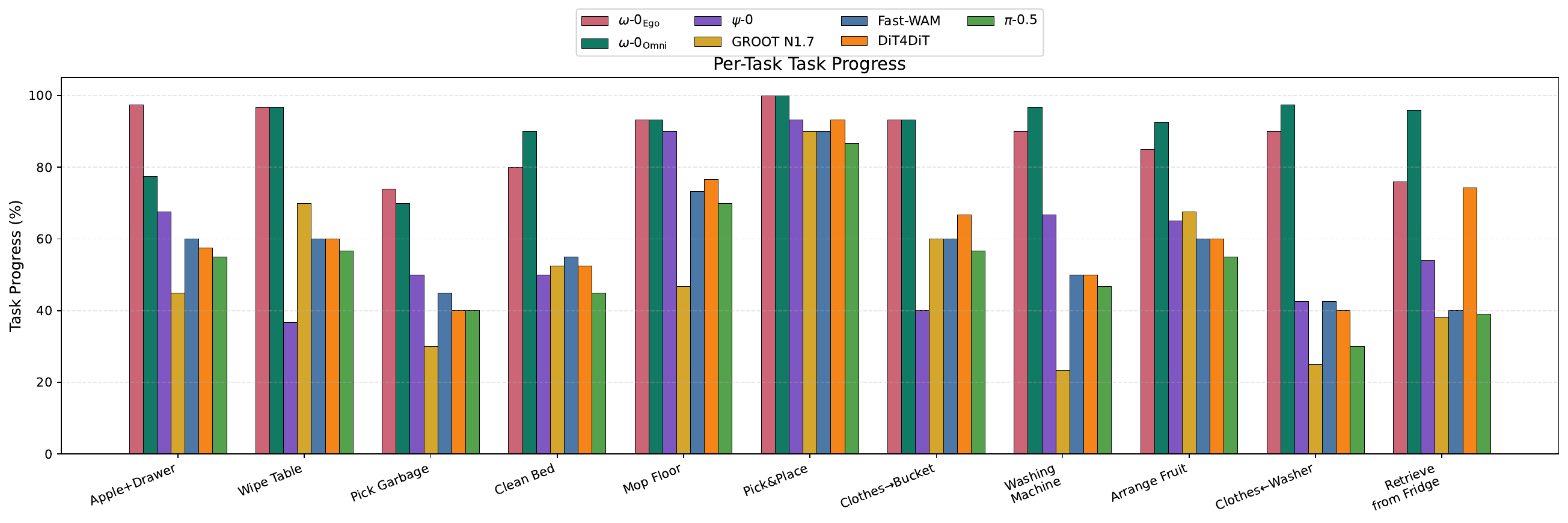}
    \caption{
    Per-task task progress comparison across seven models. 
    Task progress measures the normalized fraction of completed progress stages and provides a fine-grained view of partial task completion.
    }
    \label{fig:all7models_task_progress}
\end{figure}

Figure~\ref{fig:all7models_task_progress} reports the per-task task progress of all methods. 
Task progress measures the normalized fraction of completed progress stages, and therefore captures partial task completion even when a rollout does not achieve full success. 
This metric is particularly informative for long-horizon tasks, where a robot may complete early stages such as reaching, grasping, or opening an appliance, but fail at later stages such as precise placement or closing.

\appsubheading{app:per_task_score}{E.3 \quad Per-Task Score}

\begin{figure}[p]
    \centering
    \includegraphics[width=1.0\textwidth]{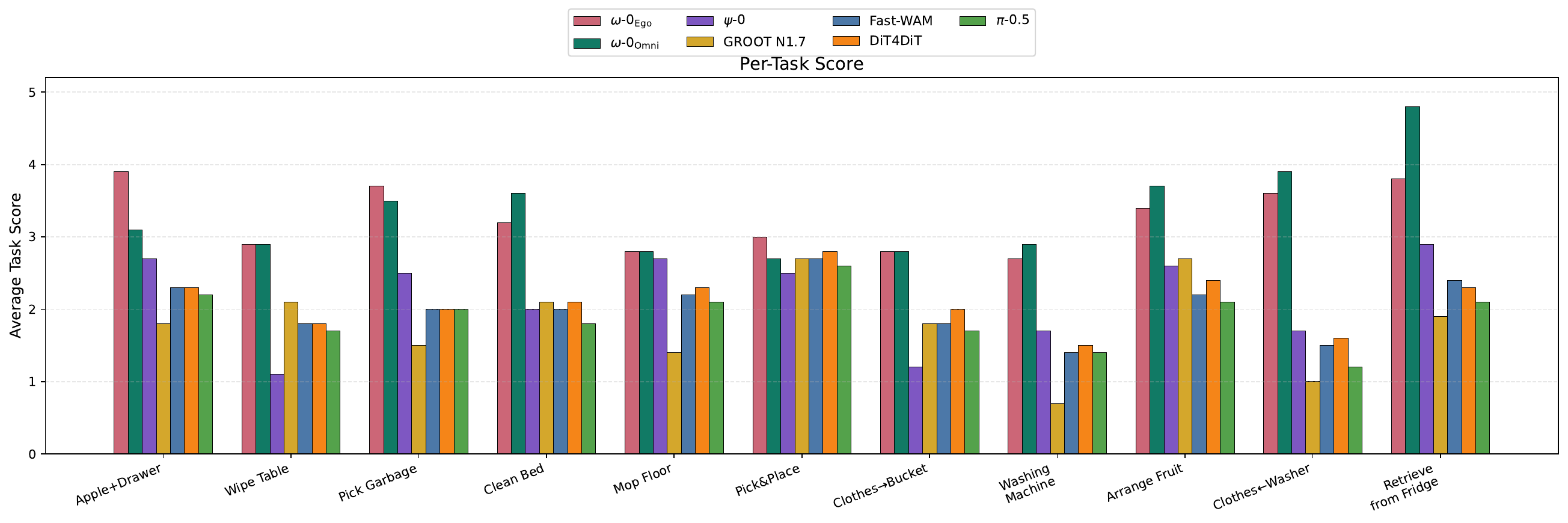}
    \caption{
    Per-task average score comparison across seven models. 
    The score denotes the average number of completed progress stages for each task.
    }
    \label{fig:all7models_score}
\end{figure}

Figure~\ref{fig:all7models_score} shows the per-task average score. 
The score is computed as the average number of completed progress stages for each task. 
Unlike success rate, which only counts fully completed trials, this score reflects how far each method progresses within a task. 
The results further confirm that $\omega$-0 improves not only final success but also intermediate task completion across diverse household scenarios.

\appheading{app:task_gallery}{D \quad Task Gallery}

\begin{figure}[b]
    \centering
    \includegraphics[width=0.9\textwidth]{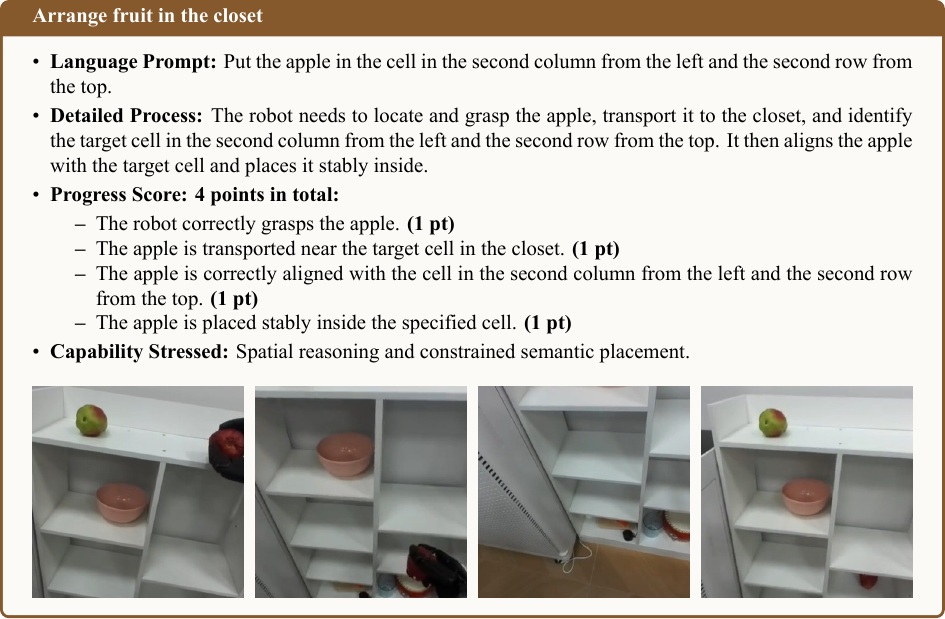}
    \caption{Task card of \texttt{arrange\_fruit\_in\_the\_closet}, which evaluates spatial arrangement and precise fruit placement in a closet.}
    \label{fig:task_arrange_fruit_in_the_closet}
\end{figure}

\begin{figure}[b]
    \centering
    \includegraphics[width=0.9\textwidth]{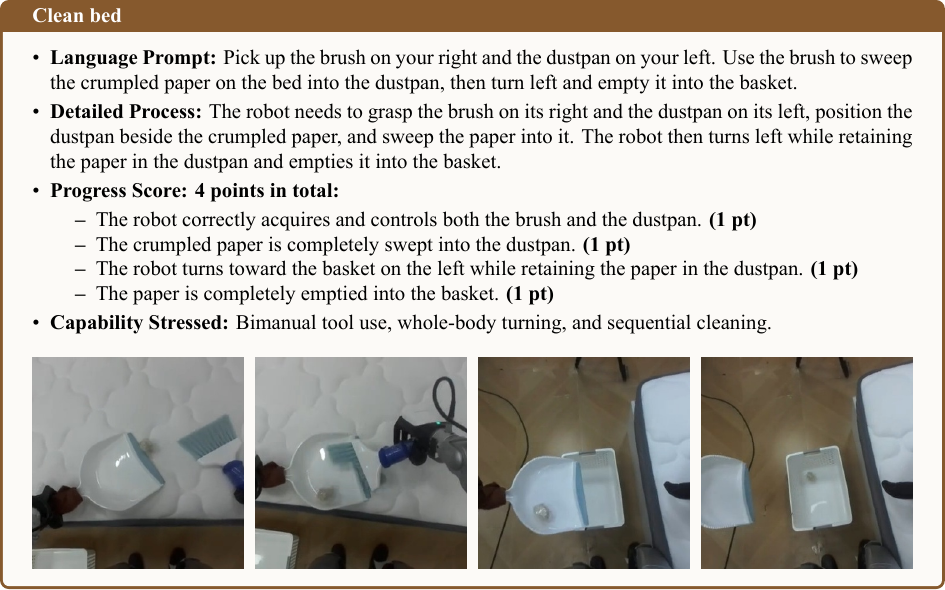}
    \caption{Task card of \texttt{clean\_bed}, which evaluates tool use, bed cleaning, object sweeping, and whole-body turning.}
    \label{fig:task_clean_bed}
\end{figure}

\begin{figure}[p]
    \centering
    \includegraphics[width=0.9\textwidth]{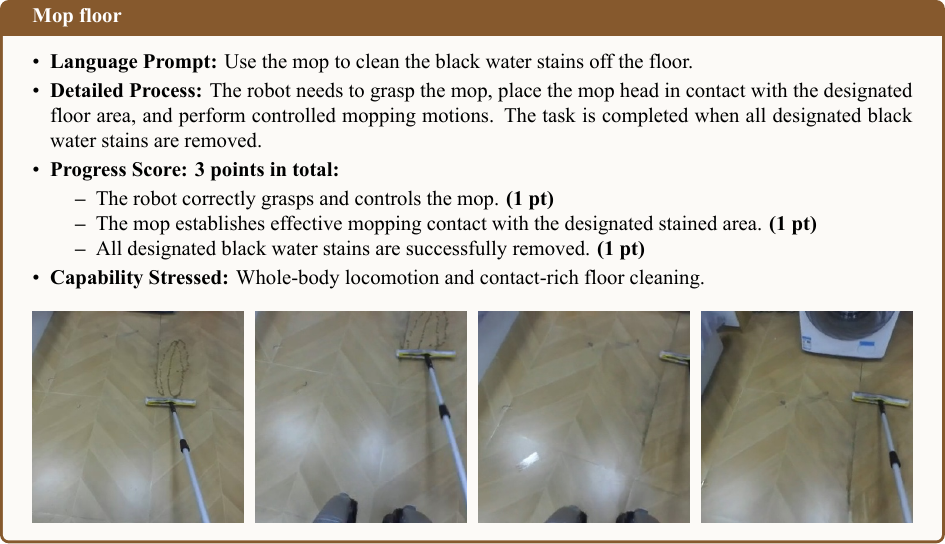}
    \caption{Task card of \texttt{mop\_floor}, which evaluates locomotion with tool use and contact-rich floor cleaning.}
    \label{fig:task_mop_floor}
\end{figure}

\begin{figure}[p]
    \centering
    \includegraphics[width=0.9\textwidth]{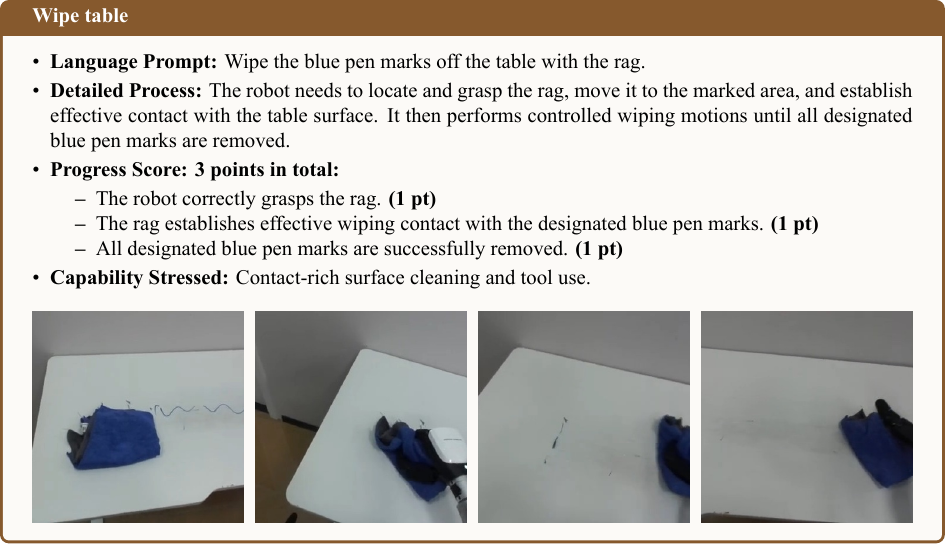}
    \caption{Task card of \texttt{wipe\_table}, which evaluates sustained contact-rich tabletop cleaning.}
    \label{fig:task_wipe_table}
\end{figure}

\begin{figure}[p]
    \centering
    \includegraphics[width=0.9\textwidth]{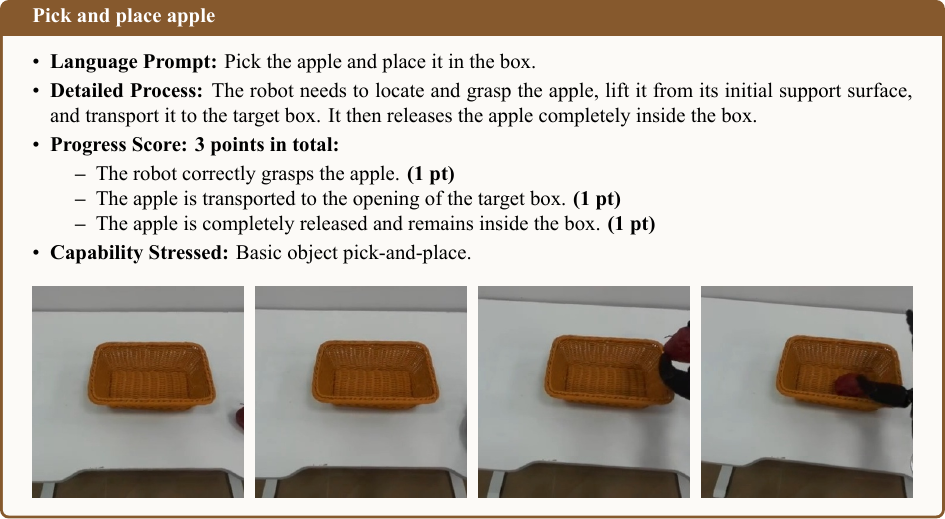}
    \caption{Task card of \texttt{pick\_and\_place\_apple}, which evaluates object grasping, transport, and placement.}
    \label{fig:task_pick_and_place_apple}
\end{figure}

\begin{figure}[p]
    \centering
    \includegraphics[width=0.9\textwidth]{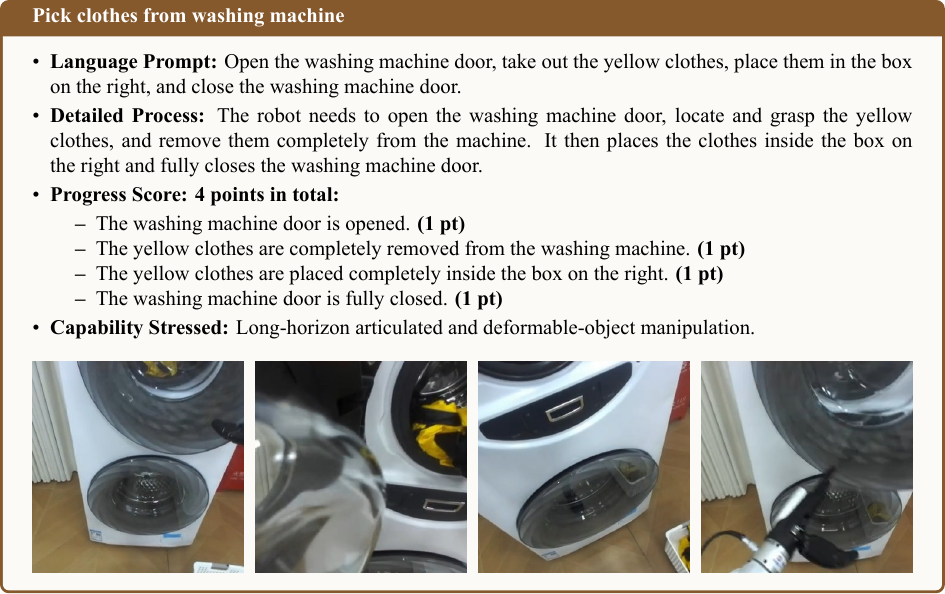}
    \caption{Task card of \texttt{pick\_clothes\_from\_washing\_machine}, which evaluates articulated-object interaction, laundry retrieval, and placement.}
    \label{fig:task_pick_clothes_from_washing_machine}
\end{figure}

\begin{figure}[p]
    \centering
    \includegraphics[width=0.9\textwidth]{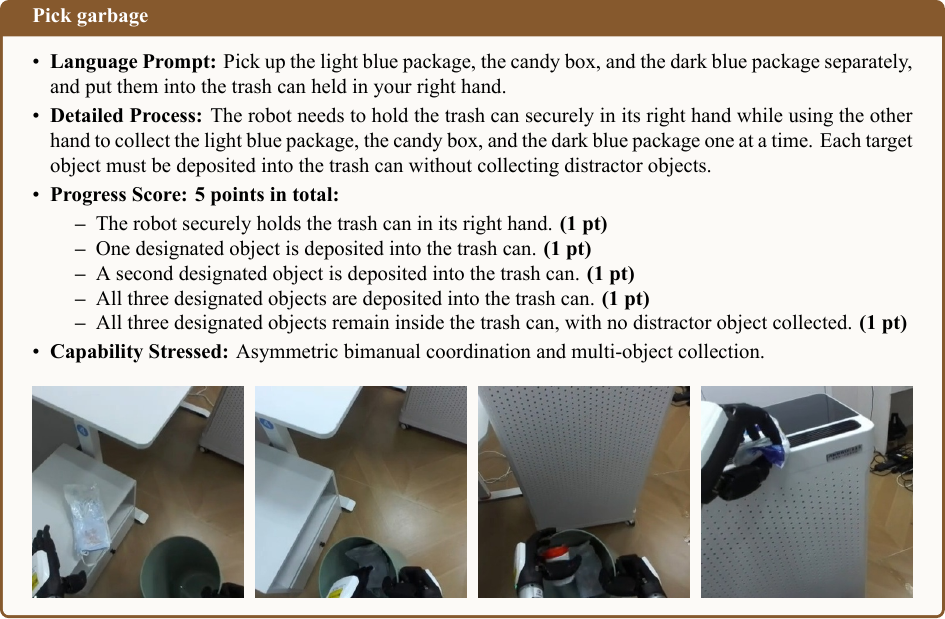}
    \caption{Task card of \texttt{pick\_garbage}, which evaluates long-horizon object collection and dual-hand coordination.}
    \label{fig:task_pick_garbage}
\end{figure}

\begin{figure}[p]
    \centering
    \includegraphics[width=0.9\textwidth]{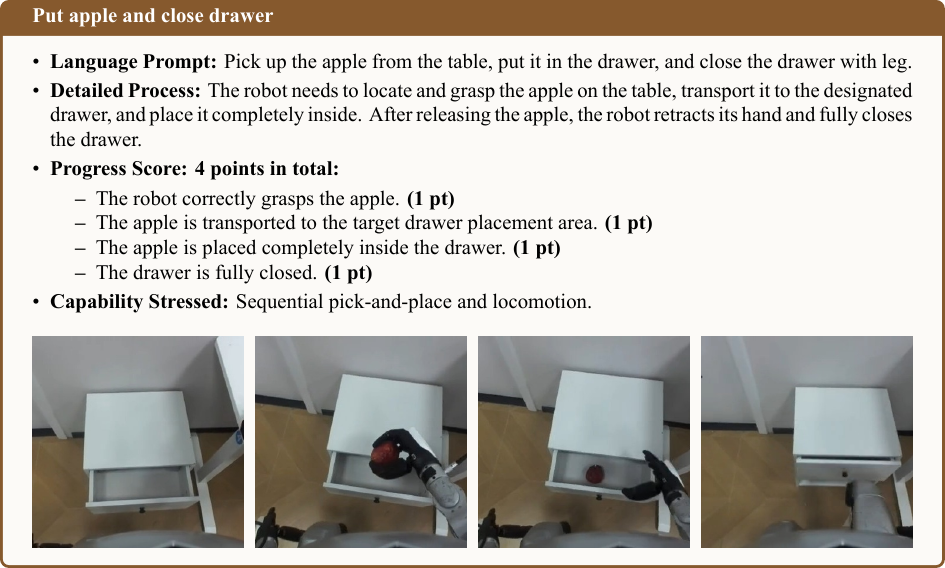}
    \caption{Task card of \texttt{put\_apple\_and\_close\_drawer}, which evaluates tabletop manipulation and drawer operation.}
    \label{fig:task_put_apple_and_close_drawer}
\end{figure}

\begin{figure}[p]
    \centering
    \includegraphics[width=0.9\textwidth]{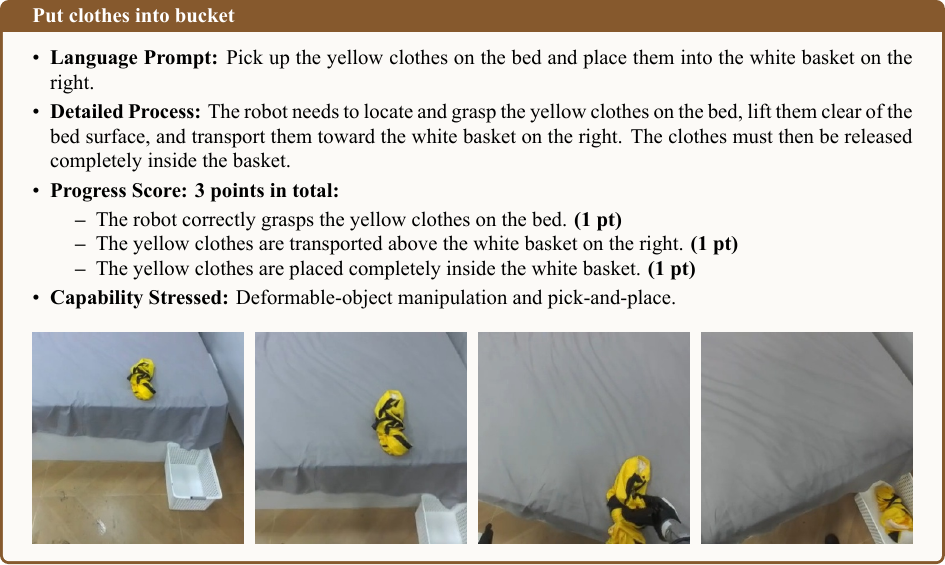}
    \caption{Task card of \texttt{put\_clothes\_into\_bucket}, which evaluates deformable-object handling and basket placement.}
    \label{fig:task_put_clothes_into_bucket}
\end{figure}

\begin{figure}[p]
    \centering
    \includegraphics[width=0.9\textwidth]{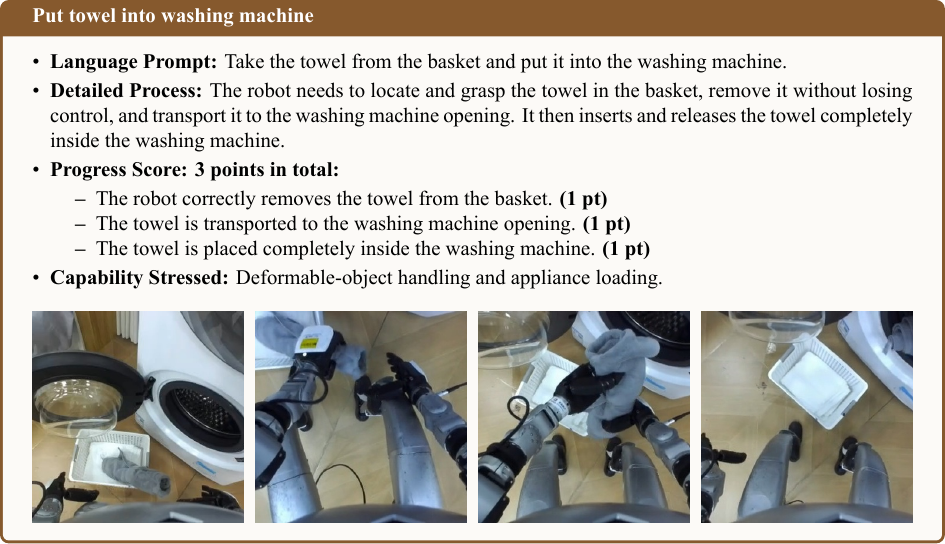}
    \caption{Task card of \texttt{put\_towel\_into\_washing\_machine}, which evaluates towel handling and appliance loading.}
    \label{fig:task_put_towel_into_washing_machine}
\end{figure}

\begin{figure}[p]
    \centering
    \includegraphics[width=0.9\textwidth]{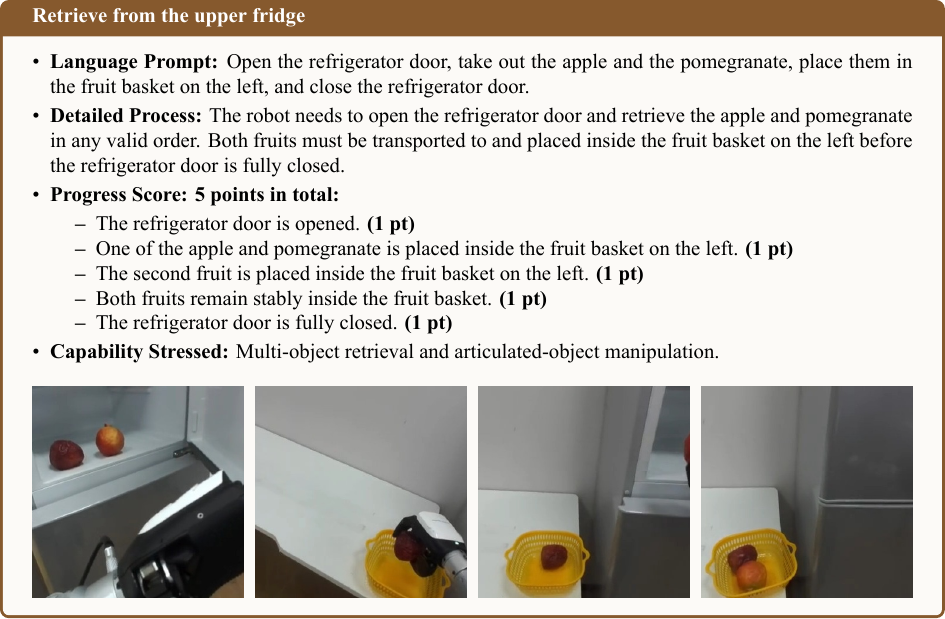}
    \caption{Task card of \texttt{retrieve\_from\_the\_upper\_fridge}, which evaluates long-horizon object retrieval and refrigerator interaction.}
    \label{fig:task_retrieve_from_the_upper_fridge}
\end{figure}

\begin{figure}[p]
    \centering
    \includegraphics[width=0.9\textwidth]{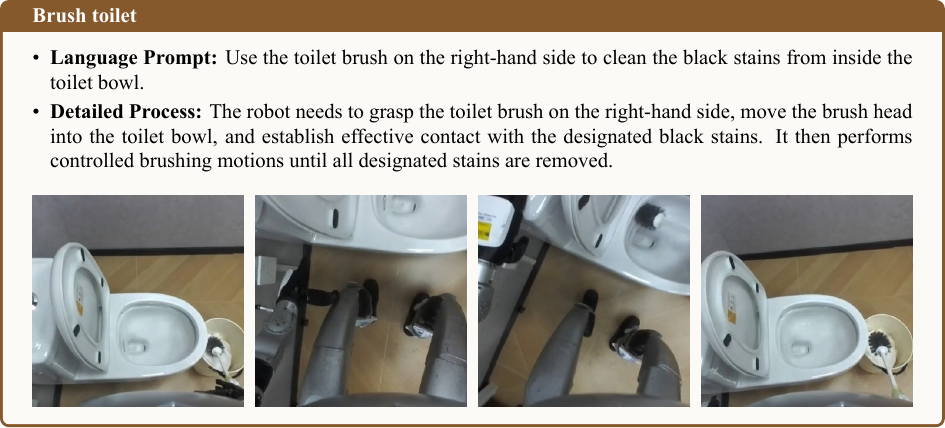}
    \caption{Task card of \texttt{brush\_toilet}, which evaluates bathroom cleaning with tool use.}
    \label{fig:task_brush_toilet}
\end{figure}

\begin{figure}[p]
    \centering
    \includegraphics[width=0.9\textwidth]{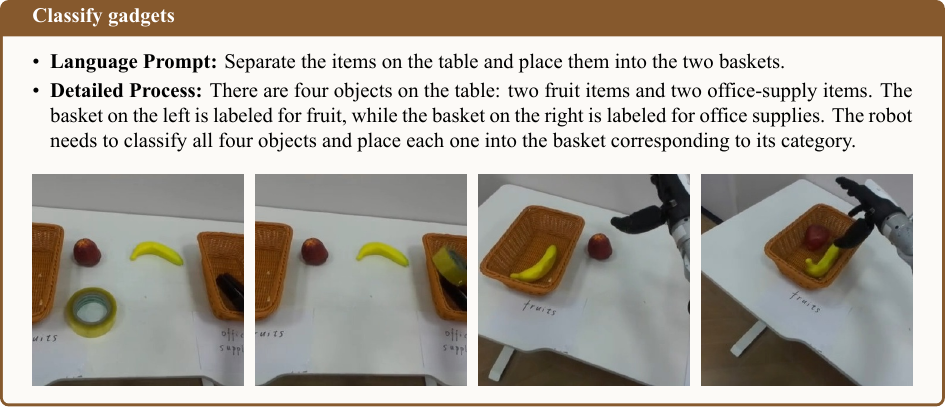}
    \caption{Task card of \texttt{classify\_gadgets}, which evaluates semantic sorting and object classification.}
    \label{fig:task_classify_gadgets}
\end{figure}

\begin{figure}[p]
    \centering
    \includegraphics[width=0.9\textwidth]{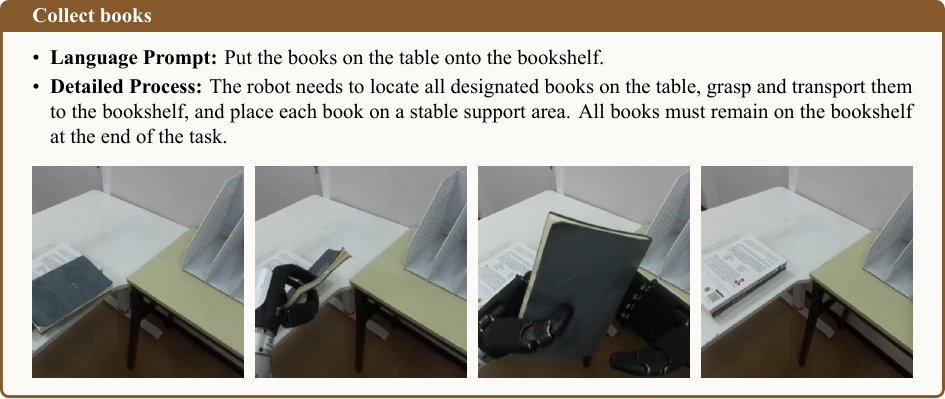}
    \caption{Task card of \texttt{collect\_books}, which evaluates object collection and tabletop organization.}
    \label{fig:task_collect_books}
\end{figure}

\begin{figure}[p]
    \centering
    \includegraphics[width=0.9\textwidth]{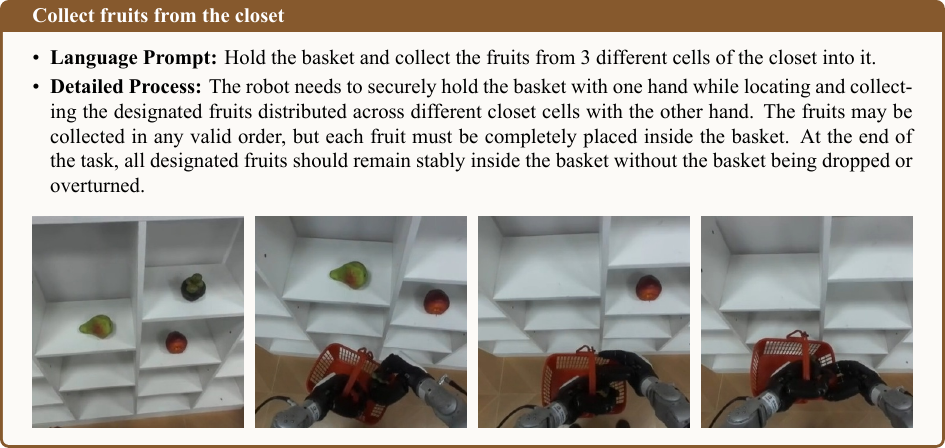}
    \caption{Task card of \texttt{collect\_fruits\_from\_the\_closet}, which evaluates fruit retrieval from structured storage.}
    \label{fig:task_collect_fruits_from_the_closet}
\end{figure}

\begin{figure}[p]
    \centering
    \includegraphics[width=0.9\textwidth]{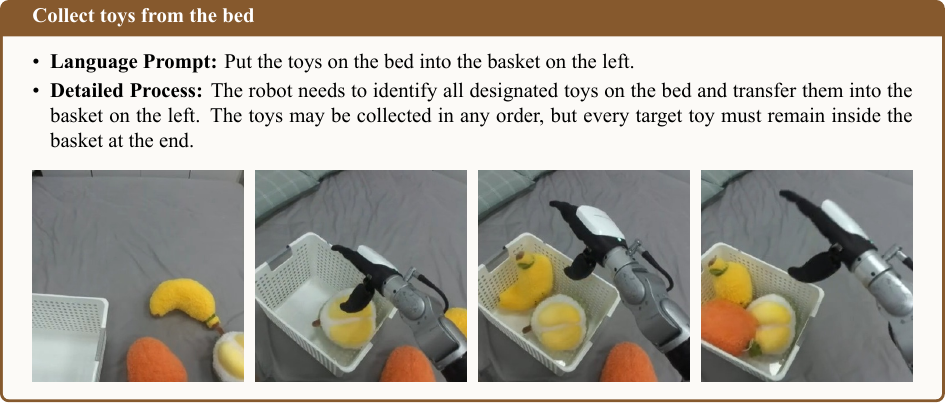}
    \caption{Task card of \texttt{collect\_toys\_from\_the\_bed}, which evaluates object collection from soft household surfaces.}
    \label{fig:task_collect_toys_from_the_bed}
\end{figure}

\begin{figure}[p]
    \centering
    \includegraphics[width=0.9\textwidth]{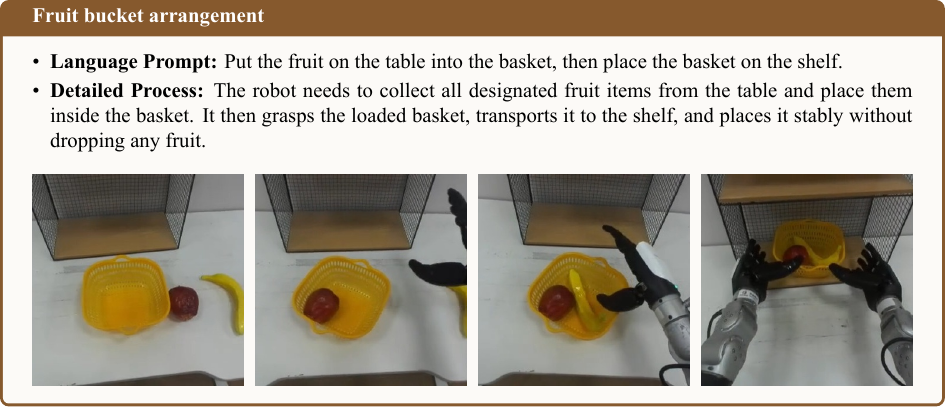}
    \caption{Task card of \texttt{fruit\_bucket\_arrangement}, which evaluates fruit arrangement and container organization.}
    \label{fig:task_fruit_bucket_arrangement}
\end{figure}

\begin{figure}[p]
    \centering
    \includegraphics[width=0.9\textwidth]{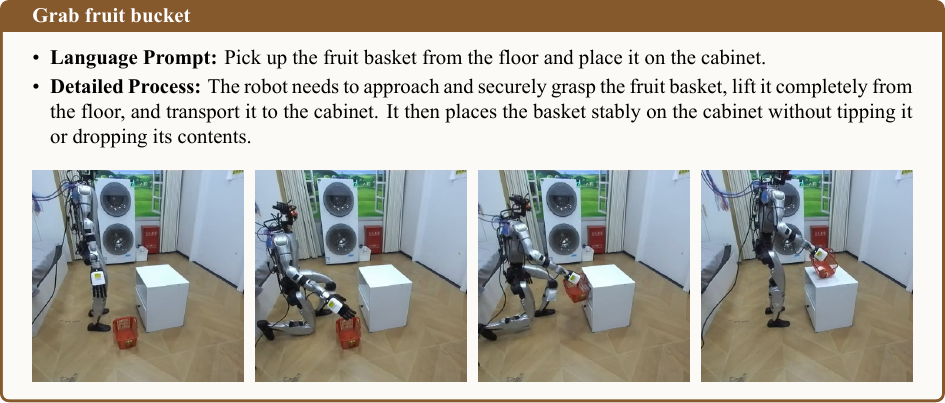}
    \caption{Task card of \texttt{grab\_fruit\_bucket}, which evaluates whole-body reaching and container grasping.}
    \label{fig:task_grab_fruit_bucket}
\end{figure}

\begin{figure}[p]
    \centering
    \includegraphics[width=0.9\textwidth]{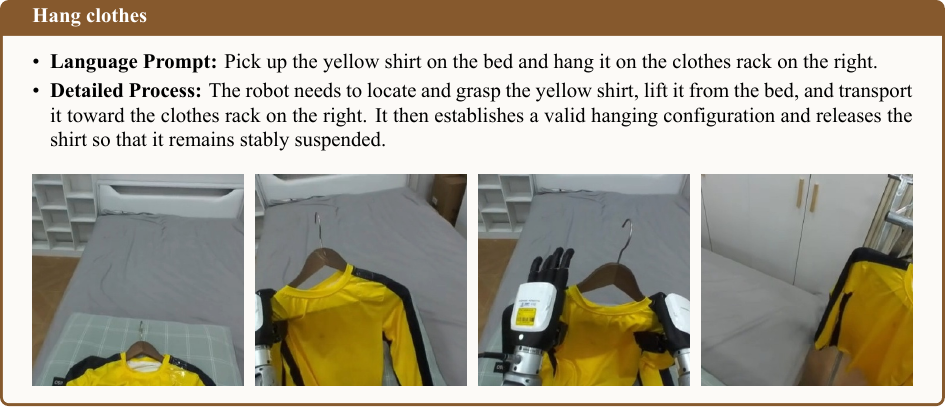}
    \caption{Task card of \texttt{hang\_clothes}, which evaluates clothes handling and hanging.}
    \label{fig:task_hang_clothes}
\end{figure}

\begin{figure}[p]
    \centering
    \includegraphics[width=0.9\textwidth]{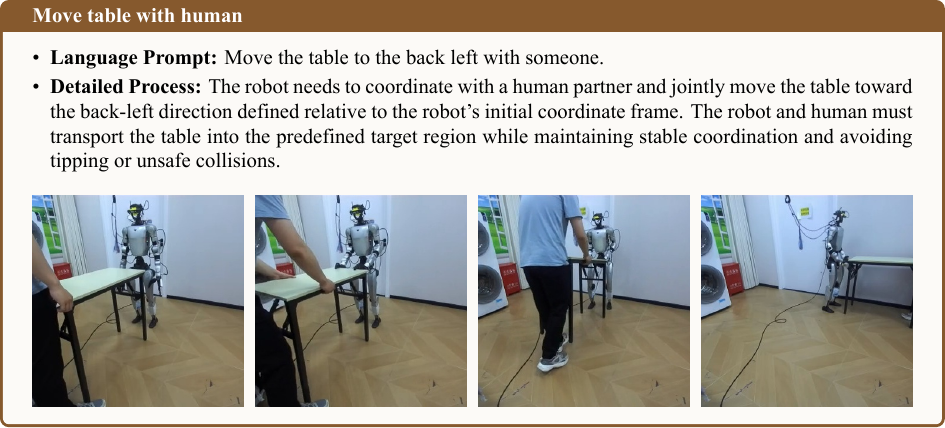}
    \caption{Task card of \texttt{move\_table\_with\_human}, which evaluates human-robot collaborative furniture moving.}
    \label{fig:task_move_table_with_human}
\end{figure}

\begin{figure}[p]
    \centering
    \includegraphics[width=0.9\textwidth]{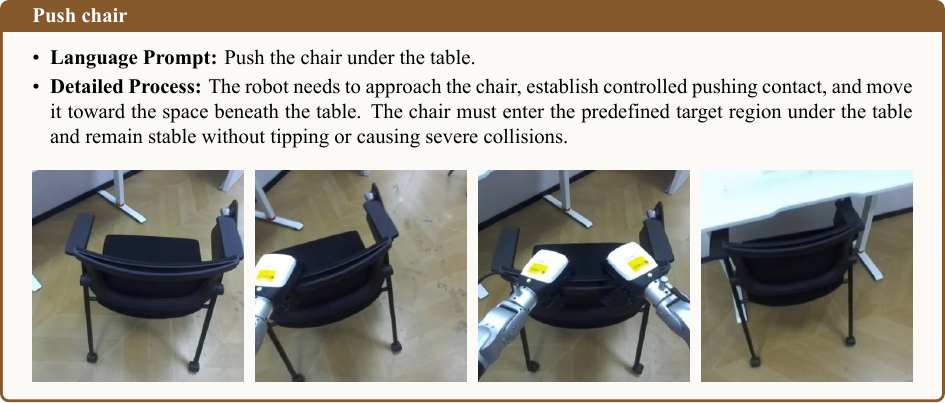}
    \caption{Task card of \texttt{push\_chair}, which evaluates furniture interaction and whole-body pushing.}
    \label{fig:task_push_chair}
\end{figure}

\begin{figure}[p]
    \centering
    \includegraphics[width=0.9\textwidth]{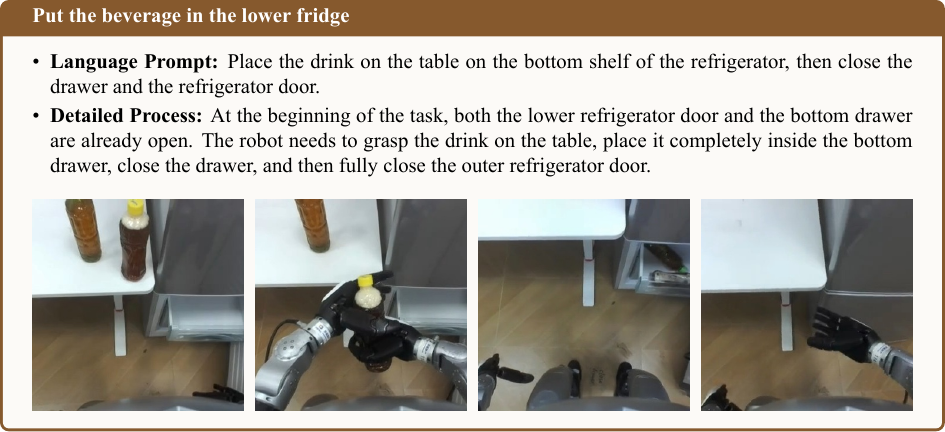}
    \caption{Task card of \texttt{put\_the\_beverage\_in\_the\_lower\_fridge}, which evaluates beverage placement and lower-fridge interaction.}
    \label{fig:task_put_the_beverage_in_the_lower_fridge}
\end{figure}

\begin{figure}[p]
    \centering
    \includegraphics[width=0.9\textwidth]{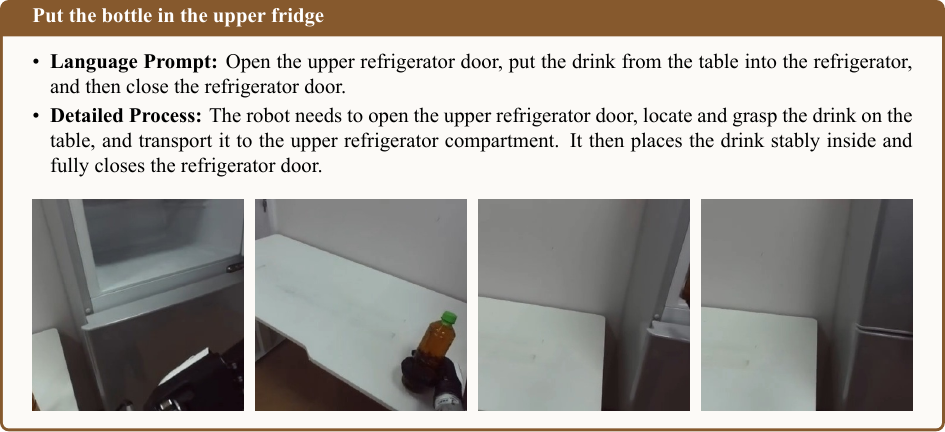}
    \caption{Task card of \texttt{put\_the\_bottle\_in\_the\_upper\_fridge}, which evaluates bottle placement into an upper refrigerator compartment.}
    \label{fig:task_put_the_bottle_in_the_upper_fridge}
\end{figure}

\begin{figure}[p]
    \centering
    \includegraphics[width=1.0\textwidth]{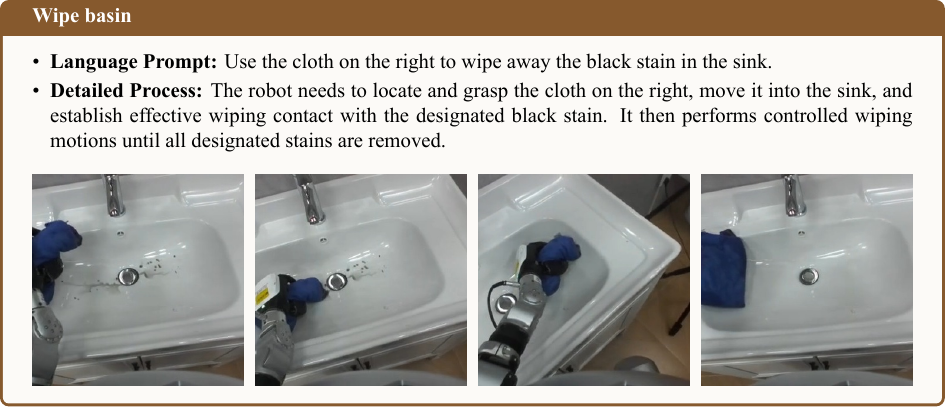}
    \caption{Task card of \texttt{wipe\_basin}, which evaluates bathroom surface cleaning.}
    \label{fig:task_wipe_basin}
\end{figure}

We provide task cards for the real-world evaluation tasks. 
Each task card shows the task scene, target objects, and representative execution sequence in Figure ~\ref{fig:task_arrange_fruit_in_the_closet}--\ref{fig:task_wipe_basin}. The evaluation tasks cover diverse household humanoid loco-manipulation capabilities, including tabletop manipulation, contact-rich cleaning, tool use, articulated-object interaction, laundry handling, object retrieval, and long-horizon mobile manipulation.

\FloatBarrier

\appsubheading{app:full_task_cards}{D.2 \quad Full $\omega$-HOME Task Cards}

We further provide task cards from the full $\omega$-HOME collection suite. 
These task cards illustrate the diversity of collected household behaviors, including bathroom cleaning, semantic sorting, object collection, furniture interaction, refrigerator operation, clothes handling, and human-robot collaborative manipulation.

\FloatBarrier

\end{document}